\documentclass[longpaper]{clv3}

\NewCommandCopy{\cnumdef}{\numdef}
\NewCommandCopy{\endcnumdef}{\endnumdef}
\let\numdef\relax \let\endnumdef\relax

\usepackage{amssymb}

\usepackage{graphicx}

\usepackage{enumitem}

\usepackage[artemisia]{textgreek}

\usepackage{booktabs}
\usepackage{multirow}

\usepackage{amsmath}

\usepackage{xcolor}

\usepackage{todonotes}
\setuptodonotes{inline}

\usepackage{xspace}

\usepackage{subcaption}

\usepackage{tikz}

\usepackage{acro}

\usepackage{wrapfig}

\usepackage{xcolor}
\usepackage{hyperref}
\definecolor{darkblue}{rgb}{0, 0, 0.5}
\hypersetup{colorlinks=true,citecolor=darkblue, linkcolor=darkblue, urlcolor=darkblue}

\usepackage[capitalise]{cleveref}

\makeatletter
\newcommand{\crefnames}[3]{%
    \@for\next:=#1\do{%
        \expandafter\crefname\expandafter{\next}{#2}{#3}%
    }%
}
\makeatother

\crefnames{part,chapter,section}{\S}{\S\S}

\newcommand{\pwc}{\textit{Papers With Code}\xspace{}}

\newcommand{\jcknumpubs}{591\xspace{}}
\newcommand{\jcknumpubnoshumanannotation}{277\xspace{}}
\newcommand{\jcknumpubalgo}{293\xspace{}}
\newcommand{\jcknumpubalgohuvalid}{16\xspace{}}
\newcommand{\jcknumpubhashumans}{314\xspace{}}
\newcommand{\jcknumpubhumanannotation}{298\xspace{}}
\newcommand{\jcknumpubsanno}{161\xspace{}}

\newcommand{\jcknumpubstp}{81\xspace{}}

\newcommand{\jcknumpubstpanno}{56\xspace{}}

\newcommand{\jcknumfeedbackloop}{68\xspace{}}
\newcommand{\jckpctfeedbackloop}{22\xspace{}}
\newcommand{\jcknumhasvalidation}{125\xspace{}}
\newcommand{\jckpcthasvalidation}{41\xspace{}}
\newcommand{\jcknumhasindirectvalidation}{25\xspace{}}
\newcommand{\jckpcthasindirectvalidation}{8\xspace{}}
\newcommand{\jcknumiterativerefinement}{31\xspace{}}
\newcommand{\jckpctiterativerefinement}{10\xspace{}}

\newcommand{\jckpctiterativefixbad}{12\xspace{}}
\newcommand{\jcknumannotatorfeedback}{18\xspace{}}
\newcommand{\jckpctannotatorfeedback}{6\xspace{}}
\newcommand{\jcknumpilotstudy}{67\xspace{}}
\newcommand{\jckpctpilotstudy}{22\xspace{}}

\newcommand{\jckpctguidelinesavailable}{60\xspace{}}
\newcommand{\jcknumannotatortraining}{55\xspace{}}
\newcommand{\jckpctannotatortraining}{18\xspace{}}

\newcommand{\jcknumqualificationfilter}{80\xspace{}}
\newcommand{\jckpctqualificationfilter}{26\xspace{}}
\newcommand{\jcknumqualificationtest}{56\xspace{}}
\newcommand{\jckpctqualificationtest}{18\xspace{}}
\newcommand{\jcknummonetaryincentive}{13\xspace{}}
\newcommand{\jckpctmonetaryincentive}{4\xspace{}}

\newcommand{\jckpctrectifyingmeasures}{41\xspace{}}
\newcommand{\jcknumfixbad}{68\xspace{}}
\newcommand{\jckpctfixbad}{22\xspace{}}
\newcommand{\jcknumdeboard}{39\xspace{}}
\newcommand{\jckpctdeboard}{13\xspace{}}
\newcommand{\jcknumagreementfilter}{29\xspace{}}
\newcommand{\jckpctagreementfilter}{9\xspace{}}
\newcommand{\jcknummanualfilter}{16\xspace{}}
\newcommand{\jckpctmanualfilter}{5\xspace{}}
\newcommand{\jcknumtimefilter}{11\xspace{}}
\newcommand{\jckpcttimefilter}{3\xspace{}}
\newcommand{\jcknumdatafilter}{46\xspace{}}
\newcommand{\jckpctdatafilter}{15\xspace{}}
\newcommand{\jcknumexpertfeedback}{24\xspace{}}
\newcommand{\jckpctexpertfeedback}{8\xspace{}}
\newcommand{\jcknumexpertfeedbacknoexpert}{22\xspace{}}

\newcommand{\jcknumautomaticchecks}{34\xspace{}}
\newcommand{\jckpctautomaticchecks}{11\xspace{}}

\newcommand{\jckpctqualityestimation}{65\xspace{}}
\newcommand{\jcknumcontrolquestions}{28\xspace{}}
\newcommand{\jckpctcontrolquestions}{9\xspace{}}
\newcommand{\jcknumadjmajorityvoting}{68\xspace{}}
\newcommand{\jckpctadjmajorityvoting}{34\xspace{}}

\newcommand{\jckpctadjbreakties}{8\xspace{}}
\newcommand{\jcknumadjmanualcuration}{29\xspace{}}
\newcommand{\jckpctadjmanualcuration}{14\xspace{}}
\newcommand{\jcknumadjquestionmark}{92\xspace{}}
\newcommand{\jckpctadjquestionmark}{46\xspace{}}
\newcommand{\jcknumadjprobabilistic}{2\xspace{}}
\newcommand{\jckpctadjprobabilistic}{1\xspace{}}
\newcommand{\jcknumadjother}{5\xspace{}}
\newcommand{\jckpctadjother}{2\xspace{}}
\newcommand{\jcknumagreement}{156\xspace{}}
\newcommand{\jckpctagreement}{52\xspace{}}

\newcommand{\jckpctagreementlabeling}{48\xspace{}}

\newcommand{\jckpctagreementtp}{31\xspace{}}
\newcommand{\jcknumagreementnohumanannotation}{7\xspace{}}

\newcommand{\jckpctagreementpercentonly}{11\xspace{}}
\newcommand{\jcknumagreementcorrelation}{7\xspace{}}
\newcommand{\jckpctagreementcorrelation}{2\xspace{}}
\newcommand{\jcknumagreementunknownmethod}{10\xspace{}}

\newcommand{\jckpctagreementinterpprevious}{5\xspace{}}

\newcommand{\jckpctagreementcomparetoliterature}{16\xspace{}}

\newcommand{\jckpctagreementinterpcustom}{27\xspace{}}

\newcommand{\jckpctagreementinterpnone}{52\xspace{}}
\newcommand{\jcknumagreementnumreported}{288\xspace{}}
\newcommand{\jcknumagreementonlysubset}{91\xspace{}}
\newcommand{\jcknumagreementfull}{197\xspace{}}
\newcommand{\jcknumagreementsamplesizemean}{1882\xspace{}}
\newcommand{\jcknumagreementsamplesizemedian}{200\xspace{}}
\newcommand{\jcknumagreementsamplesizeeqbelowonehundred}{26\xspace{}}
\newcommand{\jckpctagreementsamplesizeeqbelowonehundred}{28\xspace{}}
\newcommand{\jcknumagreementsamplesizeeqbelowtwohundred}{47\xspace{}}
\newcommand{\jckpctagreementsamplesizeeqbelowtwohundred}{51\xspace{}}
\newcommand{\jcknumagreementmeasurescountmean}{1.33\xspace{}}
\newcommand{\jcknumagreementmeasurescountmedian}{1\xspace{}}
\newcommand{\jcknumerrorrate}{54\xspace{}}
\newcommand{\jckpcterrorrate}{18\xspace{}}
\newcommand{\jcknumerrorratemean}{8.27\xspace{}}
\newcommand{\jcknumerrorratemedian}{6.00\xspace{}}
\newcommand{\jcknumerrorratesamplesizemean}{1305.68\xspace{}}
\newcommand{\jcknumerrorratesamplesizemedian}{200.00\xspace{}}
\newcommand{\jcknumerrorratenumreported}{80\xspace{}}
\newcommand{\jcknumerrorrateonlysubset}{64\xspace{}}
\newcommand{\jcknumerrorratesaloghuvalidcount}{10\xspace{}}
\newcommand{\jcknumerrorratesaloghuvalidmin}{1.40\xspace{}}
\newcommand{\jcknumerrorratesaloghuvalidmax}{16.60\xspace{}}
\newcommand{\jcknumerrorratesaloghuvalidmean}{8.93\xspace{}}
\newcommand{\jcknumerrorratesaloghuvalidmedian}{8.55\xspace{}}
\newcommand{\jcknumnoannotationprocess}{96\xspace{}}
\newcommand{\jckpctnoannotationprocess}{32\xspace{}}
\newcommand{\jcknumnoannotatormanagement}{157\xspace{}}
\newcommand{\jckpctnoannotatormanagement}{52\xspace{}}
\newcommand{\jcknumnoqualityestimation}{102\xspace{}}
\newcommand{\jckpctnoqualityestimation}{34\xspace{}}
\newcommand{\jcknumnoqualityimprovement}{135\xspace{}}
\newcommand{\jckpctnoqualityimprovement}{45\xspace{}}

\newcommand{\jckcnumrelevant}{430\xspace{}}

\newcommand{\jckcnumintroducesds}{132\xspace{}}
\newcommand{\jckcpctintroducesds}{30\xspace{}}
\newcommand{\jckcnumds}{993\xspace{}}
\newcommand{\jckcnumuniqueds}{622\xspace{}}
\newcommand{\jckcnumhaspwc}{172\xspace{}}
\newcommand{\jckcpcthaspwc}{27\xspace{}}
\newcommand{\jckcnuminall}{49\xspace{}}

\newcommand{\jckcnuminrelevant}{30\xspace{}}

\newcommand{\jckcnummeanusagecoverage}{1.60\xspace{}}
\newcommand{\jckcnummeanusageqall}{1.96\xspace{}}
\newcommand{\jckcnummeanusageqrel}{2.13\xspace{}}

\DeclareAcronym{pwc}{
    short = PwC,
    long = Papers With Code
}

\DeclareAcronym{iaa}{
    short = IAA,
    long = inter-annotator agreement
}

\DeclareAcronym{nlp}{
    short = NLP,
    long = natural language processing
}

\DeclareAcronym{nlg}{
    short = nlg,
    long = natural language generation
}

\issue{1}{1}{2016}

\title{Analyzing Dataset Annotation\\ Quality Management in the Wild}
\runningtitle{Dataset Annotation Quality Management}
\runningauthor{Klie, Eckart de Castilho, Gurevych}

\author{Jan-Christoph Klie\thanks{Corresponding author}}
\affil{Ubiquitous Knowledge Processing Lab\\ Department of Computer Science and Hessian Center for AI (hessian.AI) \\\href{https://www.ukp.tu-darmstadt.de/}{ www.ukp.tu-darmstadt.de }}

\author{Richard Eckart de Castilho}
\affil{UKP Lab}

\author{Iryna Gurevych}
\affil{UKP Lab}

\newif\ifarxiv

\arxivfalse
\arxivtrue

\ifarxiv
\renewcommand{\footmark}{}
\renewcommand{\issue}{}
\jname{}
\jinfo{}
\else
\pageonefooter{Action editor: Nianwen Xue. Submission received: 31 August 2023; revised version received: 25 January 2024; accepted for publication: 4 March 2024.}
\fi

\begin{document}
\maketitle
\begin{abstract}
Data quality is crucial for training accurate, unbiased, and trustworthy machine learning models as well as for their correct evaluation.
Recent works, however, have shown that even popular datasets used to train and evaluate state-of-the-art models contain a non-negligible amount of erroneous annotations, biases, or artifacts.
While practices and guidelines regarding dataset creation projects exist, to our knowledge, large-scale analysis has yet to be performed on how quality management is conducted when creating natural language datasets and whether these recommendations are followed.
Therefore, we first survey and summarize recommended quality management practices for dataset creation as described in the literature and provide suggestions for applying them.
Then, we compile a corpus of $\jcknumpubs{}$ scientific publications introducing text datasets and annotate it for quality-related aspects, such as annotator management, agreement, adjudication, or data validation.
Using these annotations, we then analyze how quality management is conducted in practice.
A majority of the annotated publications apply good or excellent quality management.
However, we deem the effort of 30\% of the works as only subpar.
Our analysis also shows common errors, especially when using inter-annotator agreement and computing annotation error rates.

\end{abstract}

\section{Introduction}

Having large, high-quality annotated datasets available is essential for developing, training, evaluating, and deploying reliable machine learning models~\citep{sunRevisitingUnreasonableEffectiveness2017, benderDataStatementsNatural2018, petersTuneNotTune2019, gururanganDonStopPretraining2020, sambasivanEveryoneWantsModel2021}.
Annotated datasets are also frequently used in linguistics~\citep{haselbachApproximatingTheoreticalLinguistics2012}, language acquisition research~\citep{behrensCorporaLanguageAcquisition2008}, bioinformatics~\citep{zengSurveyNaturalLanguage2015}, healthcare~\citep{susterShortReviewEthical2017}, and the digital humanities~\citep{schreibmanCompanionDigitalHumanities2004}.
Concerning machine learning, recent work has shown, however, that even datasets widely used to train and evaluate state-of-the-art models contain non-negligible proportions of questionable labels.
For instance, the \mbox{CoNLL-2003}~\citep[][named entity recognition]{tjongkimsangIntroductionCoNLL2003Shared2003} test split has an estimated 6.1\% wrongly labeled instances~\citep{reissIdentifyingIncorrectLabels2020, wangCrossWeighTrainingNamed2019}, \mbox{ImageNet} 5.8\%~\citep[][image classification]{vasudevanWhenDoesDough2022, northcuttPervasiveLabelErrors2021} and \mbox{TACRED} 23.9\% incorrect instances~\citep[][relation extraction]{stoicaReTACREDAddressingShortcomings2021}. \mbox{GoEmotions}~\citep[][sentiment classification]{demszkyGoEmotionsDatasetFineGrained2020} is estimated to contain even up to 30\% wrong labels.\footnote{\url{https://archive.ph/jQbNM}}
Using these datasets for machine learning can ---among other issues--- lead to inaccurate estimates of model performance~\citep{reissIdentifyingIncorrectLabels2020, vasudevanWhenDoesDough2022}, generalization failure due to data bias~\citep{mccoyRightWrongReasons2019}, or decreased task performance~\citep{stoicaReTACREDAddressingShortcomings2021, vadineanuAnalysisImpactAnnotation2022}.

Recently, conversational agents and search engines based on large language models trained via instruction tuning have been widely adopted in science and society~\citep{ouyangTrainingLanguageModels2022, weiFinetunedLanguageModels2022}.
Hence, datasets used for fine-tuning must be factually correct and contain as few biases as possible for the resulting models to be accurate and trustworthy and not to cause misinformation or harm.
Benchmark datasets to evaluate their performance and rankings also need to be as accurate as possible to allow fair comparisons.

Proper quality management must be conducted throughout the dataset creation process (as depicted in \cref{sec:annotation_process}) to produce high-quality datasets.
Dataset quality is not only limited to label accuracy but also encompasses aspects such as the quality of the underlying text, annotation scheme, adhering to established practices, or standards for a task, and social or data bias.
Quality management encompasses, among others, proper data selection, choice of annotators and training, creating and improving annotation schemes and guidelines as well as annotator agreement, data validation, and error rate estimation~\citep{hovyScienceCorpusAnnotation2010, alexAgileCorpusAnnotation2010, pustejovskyNaturalLanguageAnnotation2013, monarchHumanintheLoopMachineLearning2021}.
Even though there exists an extensive body of work that discusses quality management in theory (see \cref{sec:background}), we observe that this knowledge is difficult to find and to consult, as it is scattered across many different sources and usually treated as part of the general annotation process, hence often lacking depth.
Also, to the best of our knowledge, no work as of yet has analyzed whether and how these recommendations are applied in practice.
Disseminating and analyzing quality management is especially relevant in the context of a growing number of datasets being created and released, which can exhibit the aforementioned discussed dangers of low-quality data collection.

To better understand how quality management is actually performed in practice, we first survey the literature to summarize good practices regarding quality management for dataset creation.
Based on \textit{Papers With Code}\footnote{\url{https://paperswithcode.com/datasets}}, we then collect and annotate a large set of publications (\jcknumpubs{}, of which \jcknumpubhashumans{} report human annotation or validation) that introduce new text datasets, and analyze how often and how well the different quality management methods are used.
We also analyze the coverage of \textit{Papers With Code} with regard to the ACL anthology, LDC corpora and shared tasks to validate the representativeness of our collected dataset.
Finally, we summarize our findings and provide suggestions that dataset creators can use to consult and improve their annotation process.
To the best of our knowledge, this newly annotated dataset and analysis of annotation good practices is the most extensive and detailed to date.
We answer the following research questions:

\begin{description}[noitemsep]
    \item[RQ 1] What are good practices for data annotation quality management as described in the literature and derived from actual annotation projects?
    \item[RQ 2] Compared to the previously collected good practices, which methods are actually used in practice?
    \item[RQ 3] Overall, how thorough is annotation quality management conducted in practice?
\end{description}

\noindent
Our analysis shows that while many datasets are created according to good practices, several widespread issues exist.
When using inter-annotator agreement, there is a frequent lack of actual interpretation of the agreement values.
Also, sample sizes tend to be too low to make statistically sound conclusions when computing agreement and estimating the annotation error rate.
Good practices suggested by the literature, like annotator training, pilot studies, or an iterative annotation process, are only mentioned rarely.
Another interesting finding is that most of the time, adjudication is performed via majority voting; we found only three datasets that reported using probabilistic aggregation.
Overall, we find a lack of proper reporting of how the annotation process was planned and executed, who annotated, as well as which quality management methods were used.
These issues make it more difficult to gauge the quality of datasets and can hinder reproducibility.
In summary, our contributions are:

\begin{itemize}[noitemsep]
    \item We survey the literature and compile an extensive summary of quality management methods.
    \item We analyze how quality management is done in \textbf{practice} compared to the good practices we found and recommend and point out common mistakes.
    \item Based on our findings, we provide a list of recommendations that can be used by future dataset creators to improve the quality of their datasets and to avoid common pitfalls.
\end{itemize}

\noindent In order to foster further investigation into quality management for data annotation, we will also release our code\footnote{\url{https://github.com/UKPLab/qanno}\;;\;GPL v3} to collect and analyze the dataset as well as our annotations~\footnote{\url{https://tudatalib.ulb.tu-darmstadt.de/handle/tudatalib/3939}\;;\;CC BY-NC 4.0}.
Our dataset can also be used as a reference to find papers that use specific quality management methods and serve as an example of how to apply them.

\section{Background}
\label{sec:background}

In the following section, we discuss the most relevant works dealing with the dataset creation process in general and its quality management in particular.
By quality management, we understand the overall process and measures taken to reach and maintain a desirable level of quality.
The quality management measures we found are described in detail in~\cref{sec:aqm}.

\paragraph{Dataset Creation}

Dataset creation subsumes several activities, which can be coarsely divided into three categories: \textit{annotation}, \textit{production} or \textit{evaluation}~\citep{shmueliFairPayEthical2021}.
Different quality management methods are applicable or should be used depending on the task. 
Annotation or labeling means enriching data with additional information, e.g., tags for text classification.
Production encompasses activities like writing the text for question answering, paraphrasing, or summarizing.
Evaluation means using humans to compare or assess properties like quality of previously labeled or produced instances.
These can be manually or automatically created.
While also touching on text production, this article primarily discusses annotation quality management.
We call participants in a dataset creation process still annotators, even if they only perform production.

\paragraph{Dataset Creation Good Practices}

Several books and articles have been written discussing dataset creation, especially concerning the annotation process itself.
For instance, \citet{ideHandbookLinguisticAnnotation2017} collected descriptions for a wide range of different annotation projects.
\citet{pustejovskyNaturalLanguageAnnotation2013} describe the annotation process targeted towards training a machine learning model.
However, both focus mainly on setting up the respective annotation projects, collecting data, as well as developing the annotation scheme and guidelines.
Quality management is mentioned, but -- except for inter-annotator agreement -- not discussed in depth.
\citet{hovyScienceCorpusAnnotation2010} define good practices concerning conducting linguistic annotation projects.
They emphasize the importance of proper annotator selection and training and how to evaluate the resulting dataset quality using agreement.
\citet{monarchHumanintheLoopMachineLearning2021} discusses quality management for data annotation in the greatest detail.
Their focus is predominantly on how to evaluate the quality of annotated data, from simple agreement measures over comparison with gold data to annotator-specific performance.
\citet{wynneDevelopingLinguisticCorpora2005} describe good practices when creating linguistic corpora but only mention quality as important, not how to assure it.
Similarly, \citet{rohSurveyDataCollection2021} survey the different ways to collect data, for instance, via annotation, distant or self-supervision, but only bring up quality management in a short paragraph.

Several large-scale projects were conducted to develop standards and recommendations for creating language resources.
These projects are, among others, the \textit{Expert Advisory Group on Language Engineering Standards} (EAGLES) funded by the European Union (launched in 1993) or ISO/TC 37/SC 4, a technical subcommittee within the International Organization for Standardization.
The resulting standards are either relatively challenging to find or require payment.
While searching, we did not find explicit mentions of quality management or related recommendations.

\paragraph{Quality Management in Crowdsourcing} Many works have shown that crowdworkers can annotate or create datasets with similar quality compared to experts~\citep{snowCheapFastIt2008, hovyExperimentsCrowdsourcedReannotation2014}.
Proper quality management is especially important in crowdsourcing, where the risk of unreliable workers is usually higher~\citep{hovyLearningWhomTrust2013}.
An early work describing basic quality control measures to use with Amazon's Mechanical Turk is given by~\citet{callison-burchCreatingSpeechLanguage2010}.
These include having multiple annotators for each instance or using control instances to estimate annotator quality.
\citet{danielQualityControlCrowdsourcing2019} define a taxonomy of quality for crowdsourcing and extensively describe related quality control measures.
Their survey focuses on annotator management and how it is implemented in annotation tools.
Unlike our study, they do not analyze if and how quality control measures are used in practice as reported by dataset-introducing scientific publications.
\citet{leaseQualityControlMachine2011} note that the annotation platform and tools can automate quality management in crowdsourcing to a certain degree, but manual inspection is still needed.

\paragraph{Annotation Process Analysis} \citet{sabouCorpusAnnotationCrowdsourcing2014} analyze $13$ datasets created by crowdsourcing concerning how they were collected and derive good practices from this analysis.
\citet{amideiAgreementOverratedPlea2019} analyze inter-annotator agreement in the context of natural language generation evaluation and annotate 135 publications for this.
Compared to these works, we go beyond analyzing only crowdsourced datasets, have a more detailed annotation scheme, annotate as well as analyze far more publications, and summarize quality management measures and good practices in greater detail.

\paragraph{Dataset documentation checklists} 

In the past, it has been found that datasets were often not adequately documented and were just published as-is.
Therefore, several works proposed checklists and templates that should be published alongside the dataset to remedy this issue.
These are, among others, \textit{datasheets for datasets}~\citep{gebruDatasheetsDatasets2021}, \textit{dataset nutrition labels}~\citep{hollandDatasetNutritionLabel2018}, \textit{data statements for NLP}~\citep{benderDataStatementsNatural2018},  \textit{accountability frameworks}~\citep{hutchinsonAccountabilityMachineLearning2021}, or \textit{data cards}~\citep{pushkarnaDataCardsPurposeful2022}.
Similarly, more and more machine learning and \ac{nlp} conferences have adopted and are adopting reproducibility checklists for machine learning model training.
The focus of these checklists is mostly on bias, annotator background, intended use, general data statistics, data description, data origin, or preprocessing.
\citet{kottnerGuidelinesReportingReliability2011} propose a checklist that can be used when using agreement values, which is a good start but very specific to only a single aspect of quality management.
It is designed for clinical trials and might require adaptation for use in \ac{nlp}.
We did not find any checklist explicitly targeted towards overall quality management.

To summarize, while a large body of work generally discusses the dataset creation process, the parts discussing quality management are relatively scarce, quite scattered in the literature, and not easy to find.
Therefore, we summarize the literature and provide an easily referenceable set of good practices and recommendations for the dataset-creation practitioner.
We additionally annotate a large set of dataset-introducing papers for their quality management and conduct an extensive empirical evaluation of how it is applied in practice.
To the best of our knowledge, our analysis of quality management in textual dataset publications is currently the largest and the first, while not limited to a particular area like crowdsourcing.

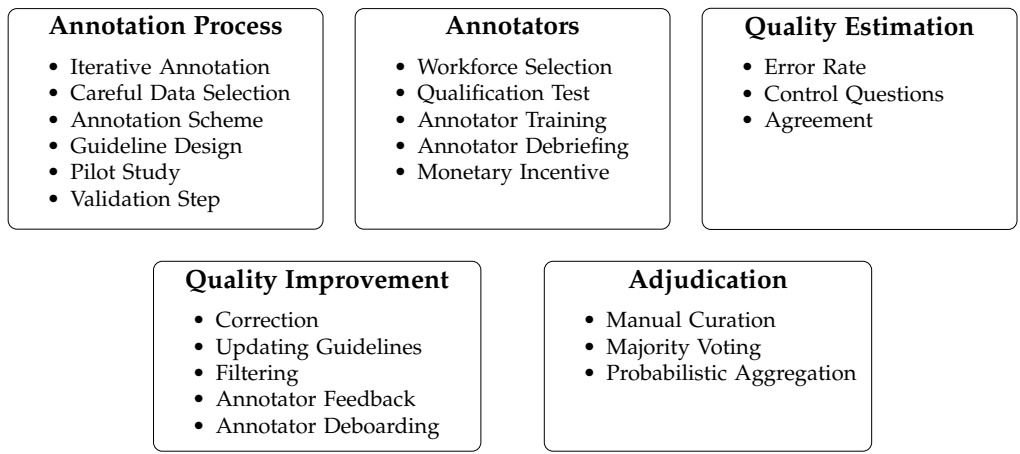
\begin{figure*}[ht]
    \centering
    \resizebox{\textwidth}{!}{
\begin{tikzpicture}[node distance=.5cm and .5cm]

 \tikzstyle{qm} = [rectangle, 
 rounded corners, 
 text width=4.8cm,
 text depth=3.cm, 
 minimum height=2.5cm, 
 align=center,
 text centered,
 draw=black]
 
 \node (annoproc) [qm] {{\large\textbf{Annotation Process}}
 \begin{itemize}[noitemsep]
     \item Iterative Annotation
     \item Careful Data Selection
     \item Annotation Scheme
     \item Guideline Design
     \item Pilot Study
     \item Validation Step    
 \end{itemize}};
 
 \node (annotators) [qm, right=of annoproc] {{\large\textbf{Annotators}}
 \begin{itemize}[noitemsep]
     \item Workforce Selection
     \item Qualification Test
     \item Annotator Training
     \item Annotator Debriefing
     \item Monetary Incentive
 \end{itemize}};
 
 \node (estim) [qm, right=of annotators] {{\large\textbf{Quality Estimation}}
 \begin{itemize}[noitemsep]
     \item Error Rate
     \item Control Questions
     \item Agreement
 \end{itemize}};
 
 \node (improv) [qm, below left=of annotators.south, text depth=2.5cm, text width=5.cm] {{\large\textbf{Quality Improvement}}
 \begin{itemize}[noitemsep]
     \item Correction
     \item Updating Guidelines
     \item Filtering
     \item Annotator Feedback
     \item Annotator Deboarding
 \end{itemize}};
 
 \node (adjudication) [qm, below right=of annotators.south, text depth=2.5cm, text width=5.cm] {{\large\textbf{Adjudication}}
 \begin{itemize}[noitemsep]
     \item Manual Curation
     \item Majority Voting
     \item Probabilistic Aggregation
 \end{itemize}};

\end{tikzpicture}}
    \caption{Quality Management methods discussed in this work. We categorize methods into annotation process, annotator management, quality estimation, quality improvement, and adjudication.}
    \label{fig:qm_method_overview}
\end{figure*}

\section{Dataset Creation Quality Management}
\label{sec:aqm}

To answer our first research question, in the following, we present the most relevant and frequently used quality management methods for dataset creation.
This list is derived from good practices stated in previous works (\cref{sec:background}) and the methods we found while surveying the dataset papers (\cref{sec:data_collection}) themselves. 
We consider the following methods \textit{good} practices for two reasons.
They are disseminated in well-acclaimed books or have been adopted by the community and are thus commonly used and tested in practice.
We thus believe that the methods discussed in the following are well-suited for managing quality.
It has to be mentioned, however, that only a few works have thoroughly investigated the exact impact of these methods on aspects like quality, time savings, or agreement (see also \cref{sec:analysis} and \cref{sec:limitations}).

Another important point to consider is to see quality management as a means towards a goal and not as a goal in itself.
Depending on the goal, for instance creating datasets with low bias, high quality, or diversity, some methods might be preferred over others.
The choice of methods should thus be based on the purpose and usage goals.

Also, when applying the ensuing methods in practice, their use can be expensive.  
Therefore, extensive quality management needs to be balanced against the annotation costs itself when working on a limited budget; a healthy compromise between the two needs to be found.

We propose a taxonomy that puts the methods into five groups related to the \textit{annotation process}, \textit{annotator management}, \textit{quality estimation}, \textit{quality improvement}, and \textit{adjudication}.
While only briefly outlining the techniques here, we refer the interested reader to each method for a more in-depth description.
An overview of the discussed methods is given in \cref{fig:qm_method_overview}.

We differentiate between two types of tasks for dataset creation (see \cref{sec:background}), namely \textit{annotation} (e.g., named entities or text classification) and \textit{text production} (e.g., writing questions and answers for question answering, paraphrasing, sentence simplification).
This distinction is important because specific quality management methods may work for one but not the other.
For example, inter-annotator agreement and adjudication are usually not applicable to text production tasks.
Both expert annotation and crowdsourcing are considered.

Our survey primarily focuses on annotation, especially label errors, but we also discuss annotation consistency, biases, and how to mitigate them.
Regarding label errors, while it is sometimes impossible to assign a single, true label due to inherent ambiguity, especially in natural language processing, deciding whether a label is incorrect is often much more straightforward. 

Before describing quality management methods, we first define what dataset quality subsumes.
Following \citet{krippendorffContentAnalysisIntroduction1980, neuendorfContentAnalysisGuidebook2016}, we suggest targeting at least the following quality aspects\footnote{Note that dataset creation projects that run over a very long time and which might be subject to external effects, such as general advances in the field or societal changes, may need other definitions for these categories or incorporate specific approaches to deal with such external effects.}:

\begin{description}[noitemsep]
    \item[Stability] A dataset creation process is stable if its output does not drift over time. Drift here means that similar phenomena are annotated similarly independent of whether they are annotated earlier or later throughout the process. Instability can, for instance, occur due to carelessness, distractions or tiredness, change in annotation guidelines, or even learning through practice.
    \item[Reproducibility] A dataset creation process is reproducible if different annotators can still deliver the same results given the same project documentation regarding process, guidelines and scheme.
    \item[Accuracy] Annotations and texts created during the process are accurate if they adhere to the guidelines and the desired outcome.
    \item[Unbiasedness] This describes the extent to which the created artifacts are free of systematic, nonrandom errors (bias).
\end{description}

\noindent
\textit{Stability}, \textit{reproducibility}, and \textit{accuracy} are also subsumed under the term \textit{reliability} in content analysis~\citep{krippendorffContentAnalysisIntroduction1980}.
\textit{Consistency} is related to \textit{stability} and \textit{reproducibility}.
\textit{Reliability} thus measures the differences that occur when repeatedly annotating the same instances; it is empirical~\citep{hardtPatternsPredictionsActions2022}.
It is required to infer \textit{validity}, that is, to show that the annotations capture the underlying phenomenon targeted~\citep{artsteinInterCoderAgreementComputational2008}, but not sufficient.
\textit{Validity} is latent and cannot be directly measured.
Therefore, proxy metrics targeting reliability, for example agreement, need to be used instead.

\subsection{Annotation Process}
\label{sec:annotation_process}

The following section describes the recommended annotation process.
It is written concerning annotation but can easily be adapted to text production as well. 

We suggest that an annotation project should start with a \textit{planning phase}.
It can encompass important preliminaries as setting the goal of data collection, making initial choices for data and annotators, setting a budget, desired quality level or reviewing the literature for similar datasets or relevant annotation practices.
Ideally, these choices are documented and become part of the dataset documentation once the dataset gets released.

The annotation scheme is often developed during an annotation project and is a living document. 
Also, as annotators only get familiar with the task during the annotation process, issues are found just then, and the data or task needs to be adapted accordingly.
Therefore, it is recommended to structure an annotation project as a sequence of cycles with iterative quality improvement actions~\cite{hovyScienceCorpusAnnotation2010, pustejovskyNaturalLanguageAnnotation2013, monarchHumanintheLoopMachineLearning2021}.
This approach is also called \textit{agile corpus creation}~\citep{alexAgileCorpusAnnotation2010}.
In each cycle, only a slice of the data is annotated: a batch. 
After the batch is annotated, it is \textit{evaluated}, and \textit{quality-improving/rectifying measures} are taken if needed.
These cycles repeat until an acceptable quality level for a sufficient number of batches has been reached.
Evaluation can be performed by inter-annotator agreement~(\cref{sec:agreement}), comparing annotations to a known gold standard to estimate annotator proficiency, or having experts or a different set of annotators inspect a subset or all instances and marking errors~(\cref{sec:manual_validation} and \cref{sec:control_instances}).

\begin{figure*}[tb]
    \centering
    \includegraphics[width=.9\textwidth]{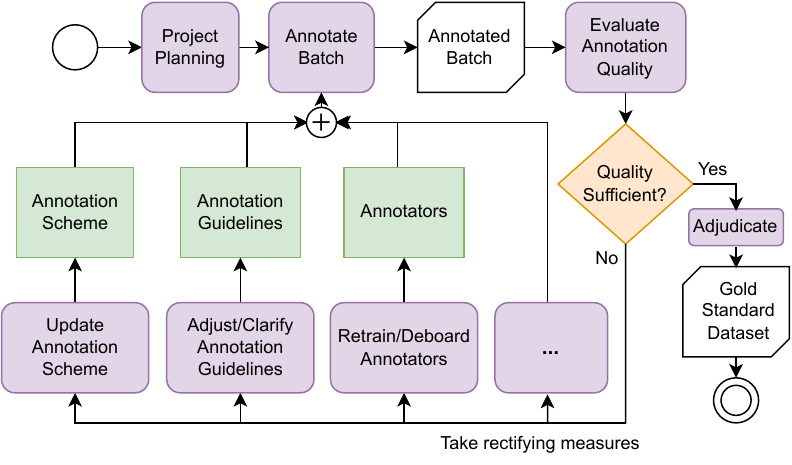}
    \caption{The recommended annotation process: After a batch of data is annotated, it is evaluated. If the quality is sufficient, it can be adjudicated. If not, several corrective measures can be taken, e.g., correcting the annotations in an additional step, annotator training, or adjusting the annotation scheme or guidelines. This is similarly applicable for text production workflows where usually no adjudication takes place.}
    \label{fig:annotation_loop}
\end{figure*}

The advantage of this iterative approach is that changes are introduced at defined points during the process.
Iterating, for example, mitigates the annotation scheme and annotations running out-of-sync and improves the chance of producing high-quality datasets.
Our take on this annotation loop is depicted in \cref{fig:annotation_loop}.
\textit{Pilot studies} are the initial iterations used to create and improve the annotation process until it is good enough for the annotation of the dataset itself.
Quality improvement measures can be, among others, retraining annotators, adjusting/clarifying the annotation guidelines or annotation scheme, onboarding or deboarding annotators, or giving back batches to annotators for correction~(cf. \cref{sec:prelim_quality_improvement}).
In later iterations of an annotation project, when the setup has stabilized, the batch sizes can be increased, and quality control can be performed less rigorously, e.g., reducing the fraction of samples inspected for quality checking or the annotations collected per instance. %
When using an iterative approach, stability of the annotation process needs to be taken into account, as changes to the process can cause differences in subsequently annotated batches.
Also, if the annotation scheme or guidelines evolve too much, then re-annotation of previously annotated material might be necessary.

\paragraph{Careful Corpus Building}

Not only are the labels assigned by the annotators important, but also the choice of texts that are annotated itself~\citep{wynneDevelopingLinguisticCorpora2005}.
Choosing texts that only rarely or even never contain the phenomena to annotate can be ineffective.
Similarly, selecting texts that are of poor quality can be detrimental and cause issues in later stages of the machine learning pipeline.
In order to achieve the best downstream task performance for trained machine learning models, texts should be representative of the data encountered in the target domain.
Hence, it is vital to check the data for errors and unwanted aspects like non-representative content or biases, ideally before it reaches the annotators.
This can be achieved by, e.g., manual inspection~\citep{bastanAuthorSentimentPrediction2020, govindarajanHelpNeedAdvice2020} (e.g., by the project manager or even as a separate preparatory annotation project) and filtering via rules~\citep{reddyCoQAConversationalQuestion2019, ghosalCICERODatasetContextualized2022} or using spell-checking and text cleaning tools~\citep{horbachInfluenceSpellingErrors2017, kimContextsensitiveSpellingCorrection2022}.

\paragraph{Annotation Scheme and Guideline Design}

The annotation scheme defines the structure, features, and tagsets of the task to annotate.
Its form and granularity can significantly impact the annotation process and the downstream machine-learning modeling.
Therefore, it must capture the information of interest.
The annotation scheme defines the annotation labels; the guidelines describe how to decide when to apply which label (e.g., disambiguating between different labels).
Properly written guidelines are essential for annotator training to achieve consistency and reproducibility, e.g., when re-annotating, extending, or creating a similar dataset on different text.
The way that the guidelines are written can by itself already introduce bias~\citep{gevaAreWeModeling2019, parmarDonBlameAnnotator2023} and therefore, great care needs to be taken when creating them.
Instead of creating guidelines from scratch for every annotation project, existing guidelines can be reused and adapted for similar settings.
In many annotation projects, the guidelines are revised several times as part of a pilot study before the actual annotation process starts~\citep{hovyScienceCorpusAnnotation2010}.

Guidelines for more complex annotation projects are often quite detailed and span many pages.
They are usually very short in crowdsourcing and often fit into the annotation screen.
Examples of excellent, extensive annotation guidelines can be found in \citet{prasadPennDiscourseTreeBank2008} or \citet{dasanmartinoNewsCategorizationFraming2022}.
For crowdsourcing, good examples are given by \citet{singhCOM2SENSECommonsenseReasoning2021} or \citet{mostafazadehGLUCOSEGeneraLizedCOntextualized2020}.

\paragraph{Pilot Study} When entering into an (iterative) annotation project, it is crucial to validate the annotation process on a smaller scale, i.e., by conducting one or more pilot studies with only a small annotator team~\citep{pustejovskyNaturalLanguageAnnotation2013}. 
Annotators in pilot studies are often the project managers themselves or a selected group of experts.
We recommend that experts or project managers conduct the initial pilot study iterations; the annotation process should then be subsequently tested with the target annotators until all questions and issues are solved.
This study should include developing the initial version of the annotation scheme and guidelines, configuring the respective annotation tooling, and developing the data pre-processing and post-processing steps~\citep{kummerfeldLargeScaleCorpusConversation2019}.
This way, issues can be spotted before investing too much effort into a flawed setup.
Ideally, the data used for pilot studies should be selected to contain as many corner cases and difficult instances as possible.
This reduces the chance that later, during the main part of the annotation project, significant adjustments need to be made that could cause costly re-annotation in case changes are not backward compatible.
The overall difficulty of the task can be gauged, and it can be tested whether experts are needed or whether well-trained contractors or crowdworkers can achieve a desirable quality level.
The expected cost can also be estimated by measuring annotation time per instance.
The feedback annotators give during this phase is essential for a well-working annotation project~\citep{monarchHumanintheLoopMachineLearning2021}.
It has to be noted, however, that if experts or project managers conduct the initial pilot study, then they may use implicit knowledge that will not transfer to more general annotators~\citep{krippendorffContentAnalysisIntroduction1980}.

\paragraph{Validation}

After an annotation step has been completed, a validation step can (and should!) be added to check whether annotations are correct and of sufficient quality.
Validation steps can take different forms based on the task and setup, e.g., experts can inspect a subset of annotations, or there can be a separate annotation phase asking for binary correctness labels.
While validation is important, it needs to be weighed against spending on annotating more instances instead if the budget is limited.

It is also possible to design a more task-dependent validation step, for which we give examples in the following.
We call this flavor of validation \textit{indirect validation}.
It is often applicable if the annotation task consists of different subtasks that depend on each other and are hence annotated sequentially. 
For question answering, a first step might be to write questions and answers. 
The validation step could then annotate which answer best fits a given question~\citep{mihaylovCanSuitArmor2018}.
For relation extraction, the first step can be marking spans and labeling their relation~\citep{yaoDocREDLargeScaleDocumentLevel2019}.
The validation step could be that annotators are given only the marked spans and are asked to label the relation.
Alternatively, the relation label could be given, and annotators are asked to mark the spans with this relation.
If annotations differ between subsequent steps, then they are potentially incorrect.
For natural language inference, the first task can be defined as writing a premise and hypothesis, given a relation (entailment, neutral, contradiction).
In the second step, the task can be to label the relation between the two given the premise and hypothesis. 
If the results in the first and second steps differ, these instances require further treatment~\citep{bowmanLargeAnnotatedCorpus2015}.

Validation is also relevant for automatically created datasets, e.g., by crawling external resources, distant or self-supervision.
It should be performed after a batch of annotations have been made and before they are adjudicated.
Validation can be part of quality estimation, which we discuss in more detail in \cref{sec:manual_validation}.

\subsection{Annotator Management}
\label{sec:prelim_annotator_management}

Dataset creation projects stand or fall by the quality of the annotators; such a project often is an exercise in people management~\citep{monarchHumanintheLoopMachineLearning2021}.
At every step, it is vital to treat annotators fairly and respectfully.
Here, we give a high-level overview of the different aspects of annotator management.
An in-depth survey of annotator management focusing on crowdsourcing is also given in~\citep{danielQualityControlCrowdsourcing2019, monarchHumanintheLoopMachineLearning2021}.
We consider both ``classic'' expert annotation and crowdsourcing in this work and point out when methods are more applicable for one or the other.

\paragraph{Workforce Selection}

The type of workforce employed considerably impacts annotation time, cost, and quality~\citep{hovyExperimentsCrowdsourcedReannotation2014}.
Which kind of annotators to employ depends, among others, on the task difficulty, availability, target language, and whether particular expertise is needed.
If the annotation task is solvable by crowdworkers, it is often an efficient way to annotate~\cite{snowCheapFastIt2008}.
For more involved tasks, trained contractors can be an alternative to hiring domain experts~\citep{chenFinQADatasetNumerical2021}.
Contractors are a middle ground between crowdworkers and experts; they are experienced in conducting annotation tasks but are not necessarily domain experts.
It is recommended to validate the workforce choice in one or more pilot studies.

\paragraph{Qualification Filter}

As a common way to filter out crowdworkers that might produce low-quality work, many crowdsourcing tools offer setting requirements for the worker.
These, for instance, can be requiring a certain percentage of accepted tasks or a certain number of already completed tasks.
\citet{kummerfeldQuantifyingAvoidingUnfair2021} analyzes the impact of these measures on quality and discusses the ethical aspects of requiring a minimum number of tasks.
They argue that it forces workers to accept a substantial amount of low-paying tasks to overcome this hurdle.
The conclusion is that there is no clear relation between quality and filtering based on the percentage of accepted, previous tasks, and number of completed tasks.
They also note that in practice, limits are often set too high.
Thus, the paper recommends either running a pilot study to get estimates for the actual requirement values or prefer qualification tests (see below) over simple filters.

\paragraph{Qualification Test}

A more elaborate way to identify good annotators is to use (paid) qualification tests~\citep{kummerfeldQuantifyingAvoidingUnfair2021}.
Before an interested annotator can participate in the primary annotation process, they must work on a small set of qualification tasks.
The answers are either compared against known answers or judged by experts.
If the performance is acceptable, then the annotator is allowed to work on the actual annotations themselves.
The difficulty of the test can be varied based on how strictly the test should filter.
For instance, task examples from the guidelines can be handed out to annotators to check whether they have been read and understood.
A more challenging test would be to use new, previously unseen tasks.
Qualification tests can and should not only be used for crowdsourcing but also when hiring contract annotators.

\paragraph{Annotator Training}

Before involving new annotators in a project, it is often helpful to train them in the annotation task at hand, to go through the guidelines with them, and make sure that everything is clear~(\citet[][p. 133]{neuendorfContentAnalysisGuidebook2016}; \citet{sabouCorpusAnnotationCrowdsourcing2014}).
Project managers and annotators can give each other feedback that can then be worked into the annotation scheme and guidelines.
Feedback is especially important if annotators find the guidelines difficult to understand or if they contain errors.
\citet{bayerlWhatDeterminesInterCoder2011} conduct a meta-study and analyze, among other aspects, the effect of training on agreement.
They show that the better and more intensely annotators are trained, the higher the agreement becomes.
Also, they point out that training is beneficial not only to crowdworkers but also to experts, as the latter might be familiar with the domain but not with the project setup at hand.
Training is also essential for annotation stability, as early in the process, annotators are often unsure and unfamiliar with the annotation process.
This changes with more time spent annotating, rendering earlier annotations potentially inconsistent with later ones.

\paragraph{Annotator Debriefing}

During and after the run of an annotation project, it is often helpful to ask one's annotators for feedback about the annotation project~\citep[][p. 134]{neuendorfContentAnalysisGuidebook2016}.
This feedback can then be used to improve the guidelines, update the annotation scheme, or alleviate issues that only became apparent while annotating.
For instance, usability issues of the annotation editor, ways to make annotation faster, or data quality issues can be spotted and fixed before it is too late.

\paragraph{Monetary Incentive}

Giving annotators additional monetary compensation in addition to their base pay might be an option~\citep{harrisYoureHiredExamination2011, hoIncentivizingHighQuality2015}.
The amount, for instance, can be based on their performance on control questions or after feedback rounds have shown that they reach the target for a bonus.
Another way is to pay annotators more for sticking to a task~\citep{parrishDoesPuttingLinguist2021}.
If monetary incentives are used, it is essential to be transparent about it, communicate the requirements beforehand, be fair, and not change the rules post-hoc. 
Also, one needs to be careful that the targets for which monetary incentives are promised are not gamed with detrimental effect towards annotation quality.~\footnote{This is also known as Goodhart's law: \textit{``When a measure becomes a target, it ceases to be a good measure''}~\citep{goodhartProblemsMonetaryManagement1984}}

\subsection{Quality Estimation}
\label{sec:prelim_quality_estimation}

After annotations have been made, their quality should be estimated and compared to the desired quality level.
In case it is insufficient, counter-measures should be taken to improve it.

\subsubsection{Manual Inspection}
\label{sec:manual_validation}

In order to judge the quality of an instance dichotomously as correct or incorrect, annotators (usually, they are different from the initial annotators) or project managers can manually inspect and grade them~\citep{pustejovskyNaturalLanguageAnnotation2013}.
Validation can either be done on a subset of instances or as a complete validation step.
In addition, after the dataset has been completely annotated, its error rate can be estimated and reported because even datasets considered gold often still contain errors~\citep{northcuttPervasiveLabelErrors2021}.
The error rate is computed by dividing the number of errors found by the number of instances inspected.
Therefore, we strongly recommend inspecting a subset of instances of the final dataset, labeling their correctness, and thereby estimating the error rate.
The notion of what is correct/of sufficient quality or incorrect/insufficient depends on the task at hand.
Hence, manual inspection is not only applicable to annotation tasks but also to text production.
There, it can be determined whether the produced instance is of sufficient quality.
For ambiguous instances in annotation tasks, one would judge whether the label makes sense at all in this context.

\subsubsection{Control Instances}
\label{sec:control_instances}

In order to gauge the performance of annotators, instances can be injected into the annotation process for which the answer is known~\citep{callison-burchCreatingSpeechLanguage2010}.
These gold instances are often obtained by having experts annotate a subset beforehand.
Another way is to compare a single annotator's submissions to the others'; the performance estimate is then the deviation from the majority vote~\citep{hsuehDataQualityCrowdsourcing2009} or the agreement~\citep{monarchHumanintheLoopMachineLearning2021}.
For example, the resulting estimates can be used to retrain annotators if they annotated too many instances incorrectly, send batches created by underperforming annotators back for re-annotation, or remove annotators from the workforce.
Well-performing annotators can also be monetarily rewarded or given tasks requiring more expertise, such as task validation or manual adjudication.

\subsubsection{Agreement}

\label{sec:agreement}

A common way to quantify the reliability of annotations and annotators is to compute their \ac{iaa}~\citep{ebelEstimationReliabilityRatings1951, krippendorffContentAnalysisIntroduction1980, krippendorffReliabilityContentAnalysis2004}.
For \ac{nlp}, it has been increasingly adopted after \citet{carlettaAssessingAgreementClassification1996} introduced agreement, coming from the field of content analysis, as an alternative to previously used ad-hoc measures.
Here, we briefly present the most popular and recommended agreement measures.
For a more in-depth treatment of agreement and how to apply it, we refer the interested reader to the excellent works of \citet{krippendorffContentAnalysisIntroduction1980, lombardContentAnalysisMass2002, neuendorfContentAnalysisGuidebook2016, artsteinInterCoderAgreementComputational2008, monarchHumanintheLoopMachineLearning2021}.

\paragraph{Percent agreement}

This is the most straightforward agreement measure.
It considers the percentage of coded units on which two annotators have agreed.
This measure, however, suffers from several issues~\citep{krippendorffContentAnalysisIntroduction1980, krippendorffReliabilityContentAnalysis2004, artsteinInterCoderAgreementComputational2008}.
First, it yields skewed results for imbalanced datasets, similar to accuracy when evaluating classification.
Second, it does not consider when annotators assign the same label by chance, for instance, in case they randomly guess or spam.
Third, percent agreement is influenced by the size of the tagset.
Therefore, it is difficult to compare across annotation schemes.
Finally, there are only two values of percent agreement that are meaningful and intuitive, which are 0\% and 100\%.
These issues together cause percent agreement to be uninformative and difficult to interpret and compare when estimating reliability.
Therefore, the usage of percent agreement is discouraged and should especially not be the only agreement measure reported.

\paragraph{Cohen's \textkappa}

In order to remedy the issues of percent agreement, \citet{cohenCoefficientAgreementNominal1960} proposes a chance-corrected coefficient, normalized to $[-1,1]$, to measure the agreement between two annotators.
Negative values indicate disagreement, $0$ the expected chance agreement, and values greater than $0$ indicate agreement.
\textkappa{} requires that the same number of annotators annotate all instances; no entries may be missing.
Also, annotations need to be categorical. It is defined as

\begin{equation*}
    \kappa_C = \frac{p_{o}-p_{e}}{1-p_{e}}
\end{equation*}

\noindent
where $p_o$ is the observed proportionate agreement and $p_e$ the chance agreement.

\paragraph{Fleiss's \textkappa{}}

\citet{fleissMeasuringNominalScale1971} extend Scott's \textpi{}~\citep{scottReliabilityContentAnalysis1955} to multiple annotators.\footnote{Fleiss' \textkappa{} is not an extension of Cohen's \textkappa{}, as it assumes similarly to Scott's \textpi{} that the labeling distributions are the same for each annotator, which Cohen's \textkappa{} does not~\citep{artsteinInterCoderAgreementComputational2008}.}
Similarly to Cohen's \textkappa{}, each  instance needs to be labeled by the same number of annotators.
In addition, Fleiss' \textkappa{} assumes that annotators for each instance are sampled randomly, it is not suitable for settings where all annotators annotate all instances~\citep{fleissStatisticalMethodsRates2003}.
It is defined as

\begin{equation*}
    \kappa_F ={\frac  {{\bar  {P}}-{\bar  {P_{e}}}}{1-{\bar  {P_{e}}}}}
\end{equation*}

\noindent
where $\bar  {P}$ measures observed agreement as the average agreement over annotator pairs and $P_{e}$ is the expected agreement by chance.

\paragraph{Krippendorff's \textalpha}

A different way to estimate agreement has been proposed by \citet{krippendorffContentAnalysisIntroduction1980}.
It is based on the quotient of observed disagreement $D_{o}$ and chance disagreement $D_{e}$:

\begin{equation*}
    \alpha = 1-{\frac {D_{o}}{D_{e}}} .
\end{equation*}

\noindent
Compared to Fleiss's \textkappa{}, Krippendorff's \textalpha{} is more powerful and versatile: it can deal with missing annotations, supports more than two annotations per instance, and can be generalized to handle even categorical,
ordinal, hierarchical, or continuous data~\cite{hayesAnsweringCallStandard2007}.
For instance, span labeling tasks like named entity recognition or relation extraction can be evaluated using a coefficient of the {Krippendorff's unitized \textalpha{} ($\alpha{}_u$) family~\cite{krippendorffReliabilityUnitizingTextual2016}.\footnote{The $\alpha{}_u$ family currently consists of four different coefficients~\cite{krippendorffReliabilityUnitizingTextual2016}. They differ in how and whether `gaps' (unannotated units) are take into consideration, whether labels or only units are used, or whether only a subset of labels are used when computing agreement. $\alpha{}_{cu}$ is the most applicable choice of the four that ignores gaps and takes label values into account.}
Unitizing means that annotators first divide the instances into smaller units and only then assign labels~\citep[][Chapter 4]{lombardContentAnalysisMass2002}.
In the context of named entity annotation, unitizing, for instance, can be marking spans that contain entities or, for object detection, drawing bounding boxes around objects of interest.
Hence, Krippendorff's \textalpha{} can also be applied to any task with a one-to-many relation between instances and annotations of different sizes.
The amount of overlap between annotations made by different annotators is also considered by $\alpha{}_u$ when computing agreement.
While being flexible, \textalpha{} is also more complicated to implement (especially in its unitizing form), has a higher runtime, and is more challenging to interpret and to compute confidence intervals for~\citep{artsteinInterCoderAgreementComputational2008}.

\paragraph{Correlation}

For specific tasks, annotation consists of assigning scores to instances on a numerical,  continuous, or discrete rating scale or a Likert scale.
These tasks are, among others, annotating sentiment~\citep{socherRecursiveDeepModels2013}, emotions~\citep{demszkyGoEmotionsDatasetFineGrained2020}, or semantic textual similarity~\citep{cerSemEval2017TaskSemantic2017}.
Correlation measures like Pearson's $r$ (linear correlation), Spearman's \textrho{} (linear correlation of ranks), or Kendall's \texttau{} (correlation of concordant/discordant ranks) are often used to compute agreement.
However, using correlation coefficients as an agreement measure is controversial, as they measure covariation, not agreement, i.e., they measure whether variables move together, but not whether they really are similar~\citep{vanstralenMeasuringAgreementMore2012, ranganathanCommonPitfallsStatistical2017, edwardsCorrelationDoesNot2021}.
This means that two annotators with different biases when assigning scores, e.g., one annotator systematically gives overly large scores while the other systematically underscores, would still have a high correlation but low agreement.
A better alternative to the aforementioned correlation coefficients is using Intraclass Correlation (ICC)~\citep{fisherStatisticalMethodsResearch1925}, which is explicitly designed to measure agreement.
Note that there are several different formulations of ICC depending on the number of judgments per instance, whether judgments are averaged before comparison, and whether there are missing observations ~\citep{shroutIntraclassCorrelationsUses1979}.
A visual method to assess agreement between continuous variables is the Bland–Altman plot~\citep{blandStatisticalMethodsAssessing1986}.
A worked example can be found in \cref{sec:appendix_correlation}.

\paragraph{Classification Metrics}

Especially for sequence labeling tasks like named entity recognition, classification metrics like \textit{accuracy}, \textit{precision}, \textit{recall}, and \textit{$F_1$} are often used between two annotators to compute agreement~\citep{brandsenCreatingDatasetNamed2020}.
We could not find any work formally analyzing the theoretical background and implications of using these metrics as an agreement measure.
However, they seem to suffer from several issues.
First, they are only applicable as pairwise agreement; having more annotators would require averaging, which might cause information loss.
Second, they are not chance-corrected~\citep {powersEvaluationPrecisionRecall2011}.
Third, using precision and recall for computing agreement also has the downside of not being symmetric.
Given two lists of labels $a$ and $b$, the precision value of $a$ and $b$ turns into the recall when swapping its arguments: $\mbox{precision}(a,b) = \mbox{recall}(b,a)$.
Being symmetrical is essential for agreement metrics, as one annotator should not be preferred over another.
This differs from classification metrics, where one input is from the gold data, and the other is usually from model predictions.

Although it is often treated as such, agreement is no panacea; high agreement does not automatically guarantee high-quality labels.
\citet{krippendorffReliabilityContentAnalysis2004, artsteinInterCoderAgreementComputational2008} emphasize that agreement only demonstrates a reliable annotation process, which is necessary for high-quality labels but is by itself not sufficient.
Further quality management, especially manual inspection, should be applied.
Agreement also does not cover whether the annotation scheme and guidelines capture the desired phenomena.
Low agreement also does not automatically mean low-quality labels, as tasks can inherently be subjective~\citep{aroyoTruthLieCrowd2015, umaLearningDisagreementSurvey2021}, i.e., there are cases where no distinct gold label exists for an instance.

Using only a single agreement coefficient value to gauge quality is often insufficient for a reliable estimate.
Therefore, more in-depth analysis is recommended~\citep{artsteinInterCoderAgreementComputational2008}.
This can be done by manually validating the annotations (cf.~\ref{sec:manual_validation}) to get an intuition for the resulting labels and why annotators disagree.
Disagreements can be caused by differences in annotator skill,  differences in the data or its difficulty~\citep{jamisonNoiseAdditionalInformation2015},  or due to ambiguity.
Other insights can be gained by computing pairwise agreement between individual annotators or by computing agreement per label~\citep{monarchHumanintheLoopMachineLearning2021}.
These statistics may identify poorly performing annotators or particularly difficult-to-decide labels.

If the sample size is chosen too small, the resulting agreement value might have only limited explanatory power~\citep{allanSampleSizeRequirements1999, shoukriSampleSizeRequirements2004, simKappaStatisticReliability2005}.
It is therefore recommended to have large parts of the dataset annotated by multiple annotators for a representative agreement value~\citep{passonneauBenefitsModelAnnotation2014}.
Ideally, every instance should be annotated by at least two annotators to draw reliable conclusions from agreement.

Several works propose value ranges for agreement coefficients and attach a semantic meaning to them.
For instance, \citet{landisMeasurementObserverAgreement1977} give labels for certain value ranges of Cohen's \textkappa{} ($\kappa_c$), e.g.,
$0.01-0.20$	\textit{slight agreement}, $0.21-0.40$	\textit{fair agreement}, $0.41-0.60$	\textit{moderate agreement}, $0.61-0.80$	\textit{substantial agreement}, $0.81-1.00$	\textit{almost perfect agreement}.
Similarly, \citet{banerjeeKappaReviewInterrater1999} say $\kappa_c > 0.75$ indicates excellent agreement, between $0.40$ and $0.75$ as fair to good agreement, and lower indicates poor agreement.
\citet{poppingAgreementIndicesNominal1988} considers $\kappa_c$ above $0.8$ as reliable.
\citet{krippendorffReliabilityContentAnalysis2004} considers their $\alpha \geq 0.8$ as reliable (later, they stated that it is the absolute lower limit and should better be $0.9$) and $0.667 < \alpha < 0.8$ should only be used to draw tentative conclusions.
An \textalpha{} value below  $0.667$ is said to indicate that the underlying labels are unreliable.

However, it must be noted that those boundaries are arbitrary, have certain assumptions (for instance, \citet{landisMeasurementObserverAgreement1977} consider only binary classification) to the task setup, and have no theoretical foundation.
In general, choosing a target agreement level that is considered good enough is very difficult; there is no universally acceptable agreement level that is correct for every setting~\citep{bakemanDetectingSequentialPatterns1997, neuendorfContentAnalysisGuidebook2016}.
\citet{lombardContentAnalysisMass2002} find that values above $0.9$ are nearly always acceptable, greater than $0.8$ acceptable in most situations and greater than $0.7$ acceptable for exploratory studies for some indices.
\citet{artsteinInterCoderAgreementComputational2008} state that these limits work well in their experience, and datasets reported with lower agreement values tend to be unreliable.
The threshold may also depend on the difficulty and subjectivity of the annotation task.
When stating agreement values, it is therefore essential to report boundaries and justify their value.
It is also recommended to compare agreement value to other works that annotate similar phenomena and tasks if possible.

Finally, the different agreement methods have several idiosyncrasies related to how they are computed and how they behave~\citep{zhaoAssumptionsIntercoderReliability2013,checcoLetAgreeDisagree2017}.
For instance, annotations with near-perfect percent agreement can have low Cohen's \textkappa{}.
When Krippendorf's \textalpha{} is applied to a large number of instances, then its computed chance agreement term increases while \textalpha{} reduces, thereby favoring smaller samples.
Agreement also decreases when having more annotators per instance, but this does not indicate worse quality; fewer annotators often just do not annotate the whole possible range~\citep{bayerlWhatDeterminesInterCoder2011}, and therefore, the agreement is an overestimate.
These characteristics can lead to non-intuitive behavior and render interpretation more difficult.

\subsection{Quality Improvement}
\label{sec:prelim_quality_improvement}

If the quality estimation shows that the annotation quality is insufficient, rectifying measures must be taken to improve it.

\paragraph{Manual Correction}

If the quality in a batch of annotations is too low, it can be returned to the annotators for further improvement.
Also, it can be routed to different, more experienced annotators to resolve issues in case instances are too difficult for the original annotators.

\paragraph{Updating Guidelines}

It can happen that the annotation guidelines do not cover certain phenomena in the underlying text, are ambiguous, or are difficult to understand.
Then, it might be appropriate to go back to the annotation scheme or guidelines and improve them~\citep{bareketNeuralModelingNamed2021}.
Updating the guidelines may require discarding previously created annotations or at least reviewing and updating them.
If quality estimation shows that similar categories have low agreement, then this can hint at that annotators have issues discerning between them.
One possible solution could be updating the annotation schema so that these categories are collapsed to a single label~\citep{lindahlAssessingArgumentationAnnotation2019}.

\paragraph{Data Filtering}

There are several scenarios in which already annotated instances should be prevented from making it into the final dataset.
Sometimes, certain instances are too ambiguous for which annotators then strongly disagree on a single, correct label~\citep{umaLearningDisagreementSurvey2021}.
Occasionally, annotations can be of low quality and should be removed.
A simple solution is to filter out these instances and not process them further.
The filtering can, for instance, be based on expert judgment or if there is no majority agreement~\citep{bastanAuthorSentimentPrediction2020}.
Sometimes, measuring the time it takes for annotators to process instances and filter out annotations with improbably high annotation times might also be helpful~\citep{ferracaneDidTheyAnswer2021}.

Before filtering based on agreement, the source of disagreement should be understood, and ideally, manual inspection of flagged instances should be performed.
Disagreements can for instance be visualized using confusion matrices.
Filtering instances has the potential disadvantage of reducing diversity, which should be considered.
Recent work also emphasizes that disagreement is inherent to natural language~\citep{aroyoTruthLieCrowd2015} and can, for instance, be used to create a hard dataset split or even directly learn from them~\citep{checcoLetAgreeDisagree2017, umaLearningDisagreementSurvey2021}.
Improving the annotation guidelines to incorporate edge cases should therefore be preferred over filtering.

\paragraph{Annotator Training through Feedback}

After annotators complete a batch, experts can manually inspect the data and give annotators feedback.
Thereby, common errors can be pointed out, and aspects to improve can be discussed~\citep{ghosalCICERODatasetContextualized2022, kirkHatemojiTestSuite2022}.
More detailed and extensive feedback might be more feasible for smaller annotator pools, e.g., contractors or expert annotators.

\paragraph{Annotator Deboarding}

If certain annotators repeatedly deliver low-quality work, removing them from the annotator team might be desirable.
One way to find these annotators is via annotation noise~\citep{hsuehDataQualityCrowdsourcing2009}, which describes the deviation of each annotator from the majority.
Another is a manual inspection by the dataset creators or more seasoned annotators.
Spammers can also be detected during adjudication (\cref{sec:prelim_adjudication}), for instance, by using MACE~\citep[][multi-annotator competence estimation]{hovyLearningWhomTrust2013}.
After deboarding annotators, it is recommended that their annotations are marked to be redone.
Even though some platforms like Amazon Mechanical Turk make it possible to withhold payment, they should still be paid for the work already done unless there is compelling evidence for excessive fraudulent behavior.

\paragraph{Automatic Annotation Error Detection (and Correction)}

Instead of having human annotators manually inspect instances and search for errors, automatic approaches can be used.
For some error types, it is possible to write checks that automatically find issues and sometimes even correct them~\citep{kvetonSemiAutomaticDetection2002, qianAnnotationInconsistencyEntity2021}.
These checks can be simple rules that define wrong surface form and label combinations and are derived from the data.
For noisy text like Twitter data or crawled forum texts, spell-checking might improve the underlying text before it is given to annotators.
A more involved approach is annotation error detection, which leverages machine learning models to automatically find error candidates, which can then be given to annotators for manual inspection and an eventual correction~\citep[e.g.,][]{dickinsonDetectingInconsistenciesTreebanks2003, northcuttConfidentLearningEstimating2021, klieAnnotationErrorDetection2023}.
Automatic checks should always be validated by human annotators to not accidentally introduce new errors.

\subsection{Adjudication}
\label{sec:prelim_adjudication}

In order to increase overall annotation reliability, oftentimes, more than one label per instance is collected.
These usually need to be \textit{adjudicated}, that is, finding a consensus to create the final dataset with one single label per instance~\citep{hovyScienceCorpusAnnotation2010}.
For reproducibility, it is suggested to not only publish the adjudicated corpus but also raw annotations by the respective annotators.
Learning from individual labels is also an option, especially in tasks with considerable ambiguity and disagreement~\citep{umaLearningDisagreementSurvey2021}; then, no adjudication is used.
While being an effective way to improve reliability, collecting more than one label per instance needs to be weighed against annotating more instances when working on a limited budget.
The most common adjudication methods are described in the following.

\paragraph{Manual Adjudication}

To create a gold corpus, skilled annotators, often domain experts, manually inspect and curate each instance to a single label~\citep{bareketNeuralModelingNamed2021}.
While slow and expensive, this approach can yield high-quality data because ties can be broken and errors corrected during this inspection procedure.
Curation can be sped up with automatic tooling, for instance, by automatically merging instances for which there is no disagreement or where the disagreement is below a certain threshold.

\paragraph{Majority Voting} When using majority voting, given an instance rated by multiple annotators, its resulting label is the one that has been chosen most often.
Instances without majority label can be discarded or given to an additional annotator to break the tie.
These are often experts but can also be (experienced) crowdworkers or contractors.
In some works, supermajority voting is used.
It means that more than 50\% of annotators must agree, e.g., at most one differing label is allowed, or even a unanimous vote is required.
Majority voting is easy to implement and a strong baseline compared to the more complex methods described in the following~\citep{paunComparingBayesianModels2018}. 
But \citet{leaseQualityControlMachine2011} notes that using majority voting might drown out valid minority voices and can reduce diversity, which should be taken into account.

\paragraph{Probabilistic Aggregation}

In majority voting, it is assumed that all annotators are equally reliable as well as skilled and that errors are made uniformly at random.
This assumption does not always apply in real annotation settings, especially for crowdsourcing.
Annotators can be better or worse in certain aspects, might be biased, spamming, or even adversarial~\citep{passonneauBenefitsModelAnnotation2014}.
To alleviate these issues, \citet{dawidMaximumLikelihoodEstimation1979} propose a probabilistic graphical model (that is referred to as Dawid-Skene, named after its inventors) that associates a confusion matrix over label classes for each annotator, thereby modeling their proficiency and bias.
The resulting aggregation is then based on weighing labels with the respective annotator's expertise for this label.
An alternative formulation called \textit{MACE} that also models spammers is given by \citet{hovyLearningWhomTrust2013}.

It has been shown that using more sophisticated aggregation techniques can yield higher-quality gold standards~\citep{passonneauBenefitsModelAnnotation2014, paunComparingBayesianModels2018, simpsonBayesianApproachSequence2019}, but majority voting is often a strong baseline.
The works mentioned above also discuss probabilistic aggregation in more detail.

\section{Data Collection and Annotation}
\label{sec:data_collection}

\begin{figure*}[th]
    \centering
    \includegraphics[width=0.95\textwidth]{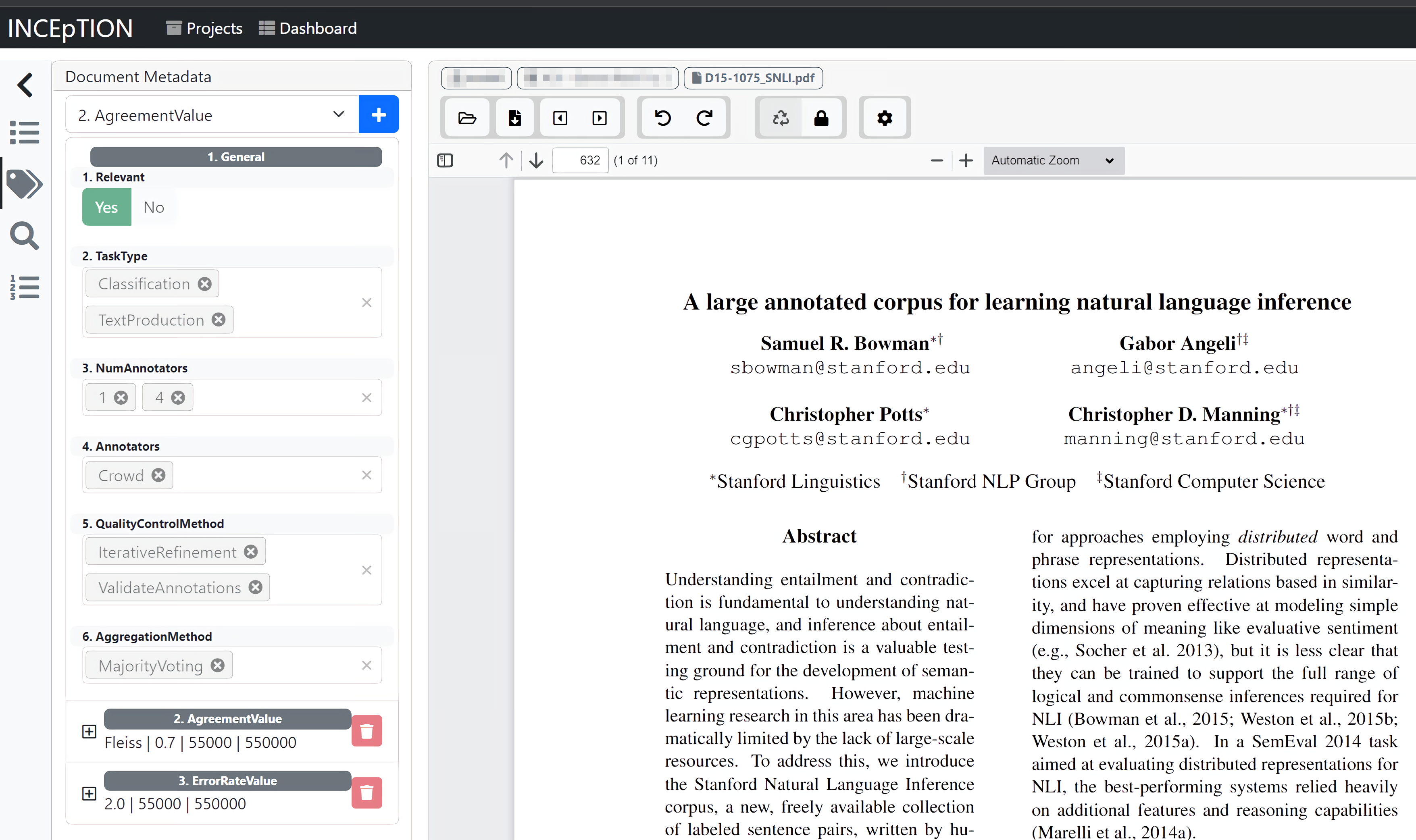}
    \caption{Annotation setup in INCEpTION. On the left, the annotation editors can be seen; on the right, a PDF viewer shows the publication to annotate directly in the browser.}
    \label{fig:annotation_editor}
\end{figure*}

\noindent To answer \textbf{RQ 2} and \textbf{RQ 3}, that is, to analyze which quality management measures are actually used when creating machine learning (research) datasets and how well works adhere to these, we collected publications that introduced new datasets and annotated them for quality aspects.

\subsection{Data Selection}
To collect relevant papers, we first attempted to use full-text search in abstracts from papers contained in the ACL anthology~\citep{gildeaACLAnthologyCurrent2018} for keywords like \texttt{dataset}, \texttt{corpus}, \texttt{treebank} or \texttt{crowdsourcing}.
This was quickly shown to be infeasible, as our search selected $13,776$ out of $36,501$ publications, showing low precision.

Instead, we chose to leverage \textit{Papers With Code}\footnote{\url{https://paperswithcode.com/}}.
This project -- among other things -- curates a list of datasets used in machine learning research with references to the publications that introduced them.
We first selected all text datasets and matched the publication title that introduced it against the ACL anthology.
We only considered papers published in top conferences as well as in their respective \textit{Findings} for the following reasons.
First, as the annotation is expensive and the budget was limited, this made the annotation more feasible by reducing the overall number of papers to read and annotate.
Second, as we are interested in collecting good practices, we hope that these publications that also passed peer review are of higher quality.
Publications from the following conferences were considered:

\begin{center}
\begin{minipage}{0.4\linewidth}
\begin{itemize}[noitemsep]
    \item AACL
    \item ACL
    \item CL
    \item COLING
    \item CoNLL
    \item EMNLP
\end{itemize}
\end{minipage}
\begin{minipage}{0.4\linewidth}
\begin{itemize}[noitemsep]
    \vspace{-1em}
    \item EACL
    \item Findings
    \item LREC
    \item NAACL
    \item TACL
\end{itemize}
\end{minipage}
\end{center}

\noindent
This yielded a total of $\jcknumpubs{}$ publications to annotate, of which $\jcknumpubhashumans{}$ mentioned human annotation or validation.
More details about our data selection and the guidelines, in particular the entire annotation scheme, including all the label values, can be found in \cref{sec:appendix_data_collection} and \cref{app:appendix_guidelines}.

\subsection{Annotation Scheme}
\label{sec:annoation_scheme}

We annotated the following aspects at document level:

\begin{description}
    \item[Manual Annotation] For our analysis, we are primarily interested in scientific publications introducing text datasets that use manual annotation in any form, which is why we annotate this aspect.
    Manual annotation may serve, e.g., for creating the labels or writing text.
    This also includes papers that only have human validation.
    \item[Task Type] There are two task types we consider, \textit{annotation} and \textit{text production}, as they require different methods for quality management.
    For instance, computing agreement is only possible for the former.
    Text production also does not lend itself to adjudication.
    \item[Number of Annotators] The number of annotators per instance whose labels are later adjudicated.
    This is only annotated for \textit{annotation} datasets, as freeform text usually is not adjudicated.
    \item[Mode of Employment] We differentiate between \emph{volunteers}, \emph{crowdworkers}, \emph{contractors} and \emph{expert annotators} (\cref{sec:prelim_annotator_management}).
    \item[Quality Management Measures] The measures mentioned in the publication to manage quality (\cref{sec:aqm}). 
    \item[Adjudication] The method of converting several annotations per instance into a single ground truth (\cref{sec:prelim_adjudication}).
    \item[Agreement] In case inter-annotator agreement was computed, we record the metric's name, the subset size if not computed on all the annotated data, and the actual value. Note that a given dataset can have more than one agreement calculation (\cref{sec:agreement}).
    \item[Error rate] In case the error rate was estimated, we record the actual value and the size of the subset that was inspected~(\cref{sec:prelim_quality_estimation}).
    \item[Overall] We assign an overall rating to each publication having human annotators based on their quality management conducted and reported. The grades are in three categories:
    \begin{description}
        \item[Excellent] Does most of the following: uses the iterative annotation process, trains annotators, computes agreement and error rate, performs extensive validation, and does human inspection throughout.
        \item[Sufficient] Uses some of the recommended techniques, but not as extensive as excellent. Has at least some validation and manual inspection.
        \item[Subpar] No agreement, validation, manual inspection, error rate, or other quality management performed and reported. The data quality, at most, relies on aggregating multiple annotations. 
    \end{description}
    We discuss limitations due to the potential subjectivity of this rating in \cref{sec:limitations}.
\end{description}

\noindent A screenshot depicting the annotation editor using this annotation scheme can be found in \cref{fig:annotation_editor}.

\subsection{Bias}
\label{sec:bias}

Using \textit{Papers With Code} as the source of publications potentially introduces several forms of bias, which we discuss in the following:

\begin{description}
    \item[Quality] As we only analyze publications from top NLP venues and for instance exclude works published in workshops, we suspect that our analysis is biased towards analyzing datasets of better quality.
    \item[Time] When looking at the distribution over publication years, we see a bias towards more recent publications.
    \item[Popularity] \textit{Papers With Code} requires volunteers to manually add datasets to the website. Therefore, the resulting collection as well as our analysis might be biased towards more popular and commonly used datasets.  
    \item[Availability] As we analyze annotation quality management by using the publication that introduced it as their proxy, we first rely on that the dataset was described in such a publication and that the publication was accepted in a top venue. Other datasets might not have been published with such an accompanying publication (this is often the case for LDC datasets), or it might have been rejected, making it unavailable for our analysis. 
    \item[Domain] As we only analyze publications from general venues and not specialized venues like workshops for narrower domains as legal or medical NLP, our collection might be biased to contain datasets that are of more general interest; particular domains might be underrepresented.
\end{description}

\noindent
In order to quantify the bias and to estimate how well \ac{pwc} covers the ACL anthology, we additionally annotated a random subset of $500$ papers from the years $2013$ to $2022$ for the datasets they use.
$2013$ as the minimal year is chosen  as older datasets are for the most part not covered by \ac{pwc} (see \cref{fig:data_stats}).
$2022$ as the maximum year was chosen as our snapshot of \ac{pwc} is from the 26th of November, 2022 (see \cref{sec:appendix_data_collection}).
Again, we limited ourselves to the aforementioned top conferences and sample $50$ papers per year randomly, resulting in $500$ papers total.
We annotated for two aspects: datasets used in the publication and whether a publication introduces new datasets.
Datasets were marked as not relevant if they do not contain dataset usage or use any other modality than text.
Subsequently, we deduplicated dataset mentions and linked them to \ac{pwc} in case they have an entry there. 
The coverage analysis can be found in \cref{sec:dsstats}.

\subsection{Annotation Process}

The annotation process we used was the same for both quality and coverage annotations.
It slightly deviates from our best practices due to limited time and money. 
We downloaded the full-text PDFs of the selected paper and annotated them in  \textit{{INCEpTION}}~\cite{klieINCEpTIONPlatformMachineAssisted2018}.
This annotation tool was chosen because it is free to use and supports annotating PDF documents out-of-the-box.
The annotations were created by the first author of this work, an experienced researcher in \ac{nlp} with a strong data annotation background.

We first conducted an initial pilot study to determine the aspects to annotate, followed by the annotation itself.
The tagset was iteratively extended during the annotation process.
After all papers had been annotated once, we did a second round to make the annotations more consistent with the now complete tagset.
Finally, we did another validation round and additionally used semi-automatic checking to improve consistency and quality further.
Thus, each publication was only annotated by a single author but inspected several times to guarantee correctness and consistency.
Due to the intricate and complex annotation scheme with many aspects, the expertise needed, and the exploratory nature of the annotations, we were only able to employ a single expert annotator.
Instead, we opted for repeated validation and correction.
In total, annotation alone took over 100 hours.
While not ideal, this is a similar setup as used in previous works surveying \ac{nlp} publications~\citep{sabouCorpusAnnotationCrowdsourcing2014, amideiAgreementOverratedPlea2019, drorHitchhikerGuideTesting2018, shmueliFairPayEthical2021}.

\section{Analysis}
\label{sec:analysis}

After having annotated a large corpus of dataset introducing data, we now use it to investigate how annotation quality management is practiced quantitatively (RQ 2) and qualitatively (RQ 3).
An overview of the overall usage of each method can be found in \cref{tab:gigatable}.
Regarding recommended good practices, it must be noted that there is no way of managing the dataset creation process that guarantees high-quality results.
Nevertheless, some methods have been shown to yield better quality than others~\citet[e.g.,][]{bayerlWhatDeterminesInterCoder2011, monarchHumanintheLoopMachineLearning2021}. 
These choices of how to manage quality have to be looked at in the context of the task to annotate for and the constraints at hand, for instance, concerning available budget, time constraints, annotator number, and experience.

Our analysis is based on what is explicitly reported in the publication; if it was not reported, we are unable consider it.
While this might cause our analysis to be less expressive and accurate, we see no simple way to study quality management in practice.
Also, this issue further emphasizes the importance of proper reporting, even if it is just in an appendix or supplementary material.

\begin{table*}[t]
    \centering
    \begin{tabular}{@{}llrr@{}}
    \toprule
    \textbf{Category} & \textbf{Method Name}            & \textbf{\#}                  & \textbf{\%}                  \\ \midrule
    \multirow{5}{*}{{Annotation Process}}
                      & Agile Corpus Creation           & \jcknumfeedbackloop          & \jckpctfeedbackloop          \\
                      & Pilot Study                     & \jcknumpilotstudy            & \jckpctpilotstudy            \\
                      & Validation Step                 & \jcknumhasvalidation         & \jckpcthasvalidation         \\
                      & Data Filtering                  & \jcknumdatafilter            & \jckpctdatafilter            \\
                      & None/Not specified              & \jcknumnoannotationprocess   & \jckpctnoannotationprocess   \\

    \addlinespace\multirow{6}{*}{{Annotator Management}}
                      & Qualification Filter            & \jcknumqualificationfilter   & \jckpctqualificationfilter   \\
                      & Qualification Test              & \jcknumqualificationtest     & \jckpctqualificationtest     \\
                      & Annotator Training              & \jcknumannotatortraining     & \jckpctannotatortraining     \\
                      & Annotator Debriefing            & \jcknumannotatorfeedback     & \jckpctannotatorfeedback     \\
                      & Monetary Incentive              & \jcknummonetaryincentive     & \jckpctmonetaryincentive     \\
                      & None/Not specified              & \jcknumnoannotatormanagement & \jckpctnoannotatormanagement \\

    \addlinespace\multirow{4}{*}{{Quality Estimation}}
                      & Error Rate                      & \jcknumerrorrate             & \jckpcterrorrate             \\
                      & Control Questions               & \jcknumcontrolquestions      & \jckpctcontrolquestions      \\
                      & Agreement                       & \jcknumagreement             & \jckpctagreement             \\
                      & None/Not specified              & \jcknumnoqualityestimation   & \jckpctnoqualityestimation   \\

    \addlinespace\multirow{9}{*}{{Quality Improvement}}
                      & Correction                      & \jcknumfixbad                & \jckpctfixbad                \\
                      & Scheme and Guideline Refinement & \jcknumiterativerefinement   & \jckpctiterativerefinement   \\
                      & Annotator Deboarding            & \jcknumdeboard               & \jckpctdeboard               \\
                      & Annotator Feedback              & \jcknumexpertfeedback        & \jckpctexpertfeedback        \\
                      & Agreement Filtering             & \jcknumagreementfilter       & \jckpctagreementfilter       \\
                      & Manual Filtering                & \jcknummanualfilter          & \jckpctmanualfilter          \\
                      & Time Filtering                  & \jcknumtimefilter            & \jckpcttimefilter            \\
                      & Automatic Checks                & \jcknumautomaticchecks       & \jckpctautomaticchecks       \\
                      & None/Not specified              & \jcknumnoqualityimprovement  & \jckpctnoqualityimprovement  \\

    \addlinespace\multirow{6}{*}{{Adjudication}}
                      & Manual Curation                 & \jcknumadjmanualcuration     & \jckpctadjmanualcuration     \\
                      & Majority Voting                 & \jcknumadjmajorityvoting     & \jckpctadjmajorityvoting     \\
                      & Probabilistic Aggregation       & \jcknumadjprobabilistic      & \jckpctadjprobabilistic      \\
                      & Unknown                         & \jcknumadjquestionmark       & \jckpctadjquestionmark       \\
                      & Other                           & \jcknumadjother              & \jckpctadjother              \\ \bottomrule
\end{tabular}

    \caption{Overview of how often each quality management (see also \cref{fig:qm_method_overview}) method  was used in absolute numbers (\#) and relative to all works that used manual annotation (\%). For adjudication, the denominator is the number of publications for which adjudication is applicable. Except for agreement, validation, and error rate, counts are directly computed from the \textit{Quality Management Measures} field of our dataset. For the other methods, we count it for the respective metric if there is at least one usage mentioned. Note that values are non-exclusive, as publications can make use of any combination of methods.}
    \label{tab:gigatable}
\end{table*}

\subsection{Dataset Statistics}
\label{sec:dsstats}

\begin{figure*}[tb]
    \centering
    \begin{subfigure}[b]{0.49\textwidth}
        \centering
        \includegraphics[width=.99\textwidth]{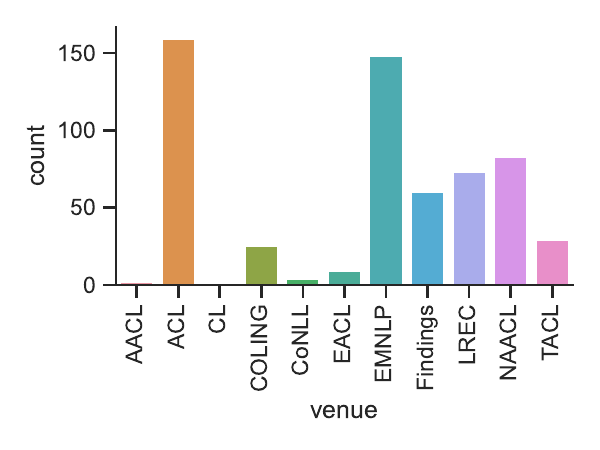}
        \caption[figure]{Publications per venue}
        \label{fig:data_stats_venues}
    \end{subfigure}
    \hfill
    \begin{subfigure}[b]{0.49\textwidth}
        \centering
        \includegraphics[width=.99\textwidth]{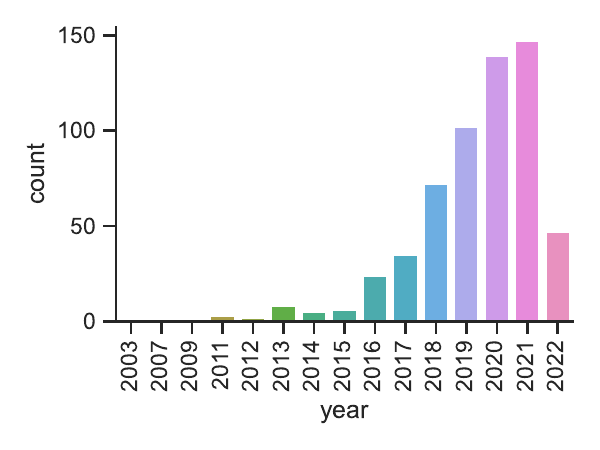}
        \caption[figure]{Publication count over time}
        \label{fig:data_stats_years}
    \end{subfigure}
    \caption{Statistics over the dataset created by annotating text dataset introducing publications obtained from \textit{Papers With Code}.}
    \label{fig:data_stats}
\end{figure*}

\paragraph{Quality Statistics}

In total, we selected and annotated $\jcknumpubs$ publications.
These were organized into three groups based on the amount of human involvement.
$\jcknumpubnoshumanannotation$ did not report any human annotation for their dataset creation.
In these cases, annotations were crawled or obtained via distant supervision or other means.
$\jcknumpubalgohuvalid$ relied on humans to validate their algorithmically created data.
$\jcknumpubhumanannotation$ had humans annotating or producing the text.
Of these $\jcknumpubhumanannotation$ publications, $\jcknumpubstp$ were introducing datasets that used annotators only for text production, $\jcknumpubsanno$ for labeling and $\jcknumpubstpanno$ for both.
Datasets that leveraged both text production and labeling were often created for tasks like natural language inference or question answering.
There, the surface forms were usually written by workers before their relationships were annotated in a follow-up step.

The resulting dataset size exceeds \citet{drorHitchhikerGuideTesting2018} who inspected $233$ papers for their analysis of statistical testing in \ac{nlp} research, as well as \citet{amideiAgreementOverratedPlea2019} who inspected $135$ publications for analyzing agreement in the context of natural language generation evaluations.
The distributions of publications per venue and over time are depicted in \cref{fig:data_stats}.
It can be seen that most were published in or after 2018. 

\paragraph{Coverage Statistics}

\acf{pwc} only contains entries for a subset of dataset-introducing publications.
To analyze the coverage and to better understand the potentially resulting bias (see \cref{sec:bias}), we conducted another annotation of $500$ papers from the anthology from the years $2013-2022$ for their dataset usage. 
Based on these annotations, we first of all can see that \jckcnumrelevant{} of publications mention relevant dataset usages.\footnote{The following metrics are with respect to relevant publications only.}
\jckcnumintroducesds{} (\jckcpctintroducesds{}\%) publications introduced new datasets of any kind.

In total, we found $\jckcnumds{}$ mentions of $\jckcnumuniqueds{}$ unique datasets, 495 datasets are only mentioned once.
Of the  $\jckcnumuniqueds{}$ unique datasets, \jckcnumhaspwc{} (\jckcpcthaspwc\%) are also contained in our dump of \ac{pwc}.
When taking our filtering of publication venues into account, we see that from the papers that we annotated for quality management, \jckcnuminall{} of all papers and \jckcnuminrelevant{} of relevant papers are  in the sample annotated for coverage as well as in the sample for quality.
In relation to our quality dataset, these make up 8\% of all and 10\% of relevant publications.

To better understand the popularity of the annotated datasets, we analyze their mention frequency.
We can see that on average, a dataset in the coverage sample was mentioned \jckcnummeanusagecoverage{} times.
In the sample for quality annotations, this was \jckcnummeanusageqall{} for all publications and \jckcnummeanusageqrel{} for only the relevant ones.
While not being a large difference, this still indicates that our sample based on \ac{pwc} is slightly biased towards more popular datasets.
 
Finally,  we find that our dataset in particular does not cover most LDC corpora or datasets introduced as part of shared tasks.
These are, for instance CoNLL, WMT, SemEval, or TAC.

\paragraph{Bias} We used  \ac{pwc} in order to reduce the effort of finding publications that introduce new datasets in the first place.
The aforementioned statistics indicate that our sampling using \ac{pwc} introduces biases towards more popular, more recent and on average, higher quality dataset.
While not ideal, we argue, however, that this is not necessarily a disadvantage, as the datasets that we analyzed are actually frequently used  in practice.
Thus, their quality has direct impact on the research community. 
Also, with being more popular, we hope that their quality management also follows good practices comparatively more often.
While having a seemingly low coverage overall, our sample size nonetheless is much larger compared to previous work, still yields interesting insights, and was already costly to annotate.

Bias in time, popularity or domain might be an issue, as there could be practices from the past that are falling through our cracks that would be relevant and interesting for the general public.
We alleviated this issue by also surveying other literature like books and by collecting and analyzing a large corpus of dataset-introducing publications.

Our analysis of annotation quality management it is still a valuable contribution, especially in combination with our survey of good practices and a good start for future work.
Also, we are interested in finding issues and to offer solutions for their alleviation, having unbiased counts is desirable but not crucial.
We hence suspect that the statistics derived overestimate quality compared to the general populace and that our analysis are potentially too positive.
The statistics that follow thus should be seen as an optimistic estimate.
Finally, it has to be noted that the the resulting dataset is a side product of the survey and should be seen in this context.
While we have taken the utmost care during annotation, the dataset is not intended to be used in machine learning or other areas where quality needs to be very high and absolute.

\subsection{Overall}

To better understand how well quality management is performed in practice (RQ 3), we assigned each work an overall score.
Their distribution is depicted in \cref{fig:overall_barchart}. 
It can be seen that around 45\% of publications perform well, and 25\% employ excellent quality management according to our annotation scheme and guidelines.
However, we also find that circa 30\% only conduct subpar quality management.
These often either did not report the annotation process at all or just very briefly and did not mention that they applied any quality management.

\begin{figure}[bht]
    \centering
    \includegraphics[width=0.7 \textwidth ]{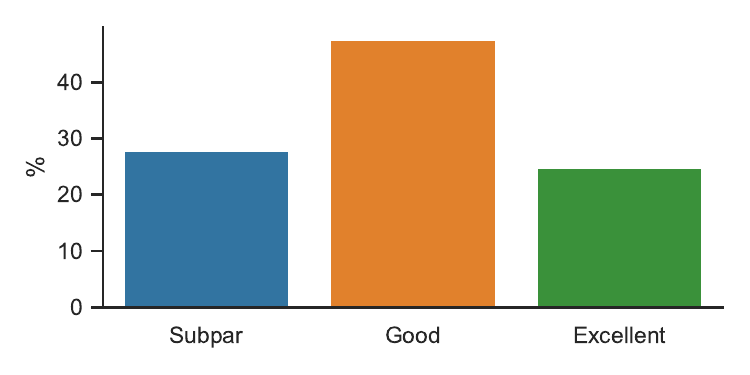}
    \caption{Distribution of percentage of papers over subjective quality management quality. Mostly, quality management was good or excellent, but a large fraction is only subpar.}
    \label{fig:overall_barchart}
\end{figure}

\subsection{Annotation Process}

In the following, we analyze the publications concerning their annotation process.

\paragraph{Annotation Scheme and Guidelines}

Of the $\jcknumpubhumanannotation$ publications having human annotators, $\jcknumfeedbackloop$ ($\jckpctfeedbackloop$\%) reported having an iterative refinement loop, which is our recommended annotation process.
This loop was mainly used for iteratively refining the annotation guidelines after doing pilot studies (\jckpctiterativerefinement \%) or repeatedly correcting instances until they reached sufficient quality (\jckpctiterativefixbad \%).
$\jcknumannotatorfeedback$\ $(\jckpctannotatorfeedback$\%) works reported that their annotators gave feedback on the task during annotation so that the annotation process could be improved.

$\jckpctguidelinesavailable$\% of publications with manual annotation described their annotation scheme, showed their annotation interface, or published their annotation guidelines together with the dataset itself in some form. 
Not reporting annotation schemes and guidelines causes several issues.
First, these cannot be checked and reviewed, making it difficult to assess their quality.
Second, not making it available is a significant obstacle to reproducibility or later extensions.
In several cases, the reader was referred to supplementary material or appendices, which we could not find in the publication or online.

\paragraph{Pilot Study}

Overall, only $\jckpctpilotstudy$\% of the publications mentioned to have conducted a pilot study.
This value is relatively low, as pilot studies are an essential tool to dial in the annotation scheme and guidelines and to get feedback from the annotators.
As we only rely on what is mentioned in publications, we cannot say whether the authors considered this method common and thus did not see the need to mention that they conducted a pilot study or that it is indeed not done often enough.

\paragraph{Validation}

In many cases, annotations were validated as an additional step in the overall process either by the annotators themselves or by having experts check them ($\jckpcthasvalidation$\%).
For automatically annotated data, only $\jcknumpubalgohuvalid$ out of $\jcknumpubalgo$ reported that they employed human validators. 
Not validating can be an issue; for example, datasets created solely by distant supervision can contain many labeling errors~\citep{mintzDistantSupervisionRelation2009}.
$\jcknumerrorratesaloghuvalidcount{}$ of these publications also reported the resulting error rate, which ranges from $\jcknumerrorratesaloghuvalidmin{}\%$ to $\jcknumerrorratesaloghuvalidmax{}\%$ with mean $\jcknumerrorratesaloghuvalidmean{}\%$ and median $\jcknumerrorratesaloghuvalidmedian{}\%$, showing the importance of validation.
We found $\jcknumhasindirectvalidation$ publications that reported indirect validation ($\jckpcthasindirectvalidation{}\%$).

\subsection{Annotator Management}

\begin{figure*}[t]
    \centering
    \begin{subfigure}[b]{0.49\textwidth}
        \centering
    \includegraphics[width=\textwidth]{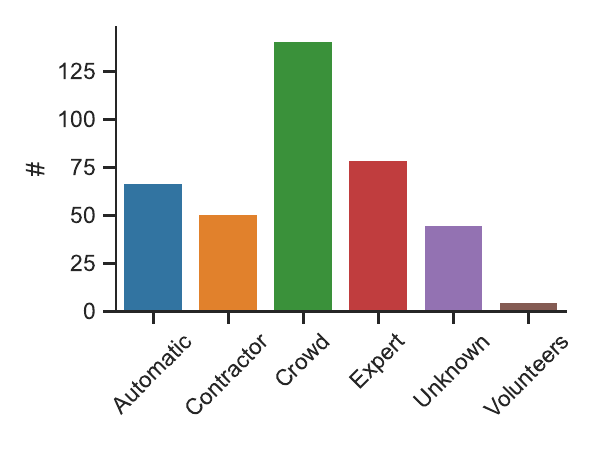}
        \caption{Absolute number of annotators by type.}
        \label{fig:anno_stats_anno}
    \end{subfigure}
    \hfill
    \begin{subfigure}[b]{0.49\textwidth}
        \centering
    \includegraphics[width=\textwidth]{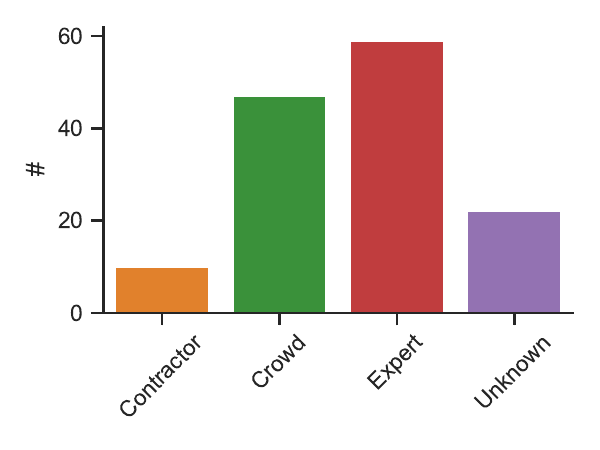}
        \caption{Absolute number of validators by type.}
        \label{fig:anno_stats_valid}
    \end{subfigure}
    \caption{Distribution over annotator types. For annotation (a), crowdsourcing is used the most; for validation (b), it is experts. Note that a publication, respectively dataset, can leverage more than one annotation type.}
    \label{fig:anno_stats}
\end{figure*}

The distribution over different annotator types is shown in \cref{fig:anno_stats}.
Overall, publications mostly used crowdworkers or experts for their annotations.
For validation, experts were more commonly selected.
In many cases, the kind of annotators used was also not reported.

We find that the preferred method to filter out annotators, especially crowdworkers, is by requiring a certain number of previous successful tasks and a high acceptance rate ($\jckpctqualificationfilter$\%).
Qualification tests, recommended by \citet{kummerfeldQuantifyingAvoidingUnfair2021} over filters, are also often employed ($\jckpctqualificationtest$\%).
Annotators are given training only in $\jckpctannotatortraining$\% of cases, which we find pretty low compared to the benefits it might give.
Out of these cases, training was overwhelmingly given to contractors and crowdworkers; only one publication mentioned that experts were trained.
We note, however, that even experts should be given training, as being an expert does not automatically indicate familiarity with the annotation setup and scheme at hand~\citep{bayerlWhatDeterminesInterCoder2011}.
Only in a few cases ($\jckpctexpertfeedback$\%) is it explicitly stated that annotators were given feedback on their work or that annotators give feedback to improve the annotation process ($\jckpctannotatorfeedback$\%).
While not being reported, we assume that training and feedback were given in many more cases, especially for contractors.
Better interaction between project leads and annotators is one reason contractors are typically chosen over crowdworkers.
$\jcknummonetaryincentive$ ($\jckpctmonetaryincentive$\%) publications mention some kind of additional monetary incentive.

\subsection{Quality Estimation}

The quality of the dataset created needs to be estimated during and after its creation so that its quality can be guaranteed and countermeasures can be taken to improve it if needed.
Overall, we find that two main techniques were used for this, which are agreement ($\jckpctagreement$\%) and error rate estimation ($\jckpcterrorrate$\%).
We analyze these in more detail in \cref{sec:analysis_error_rate} and \cref{sec:analysis_agreement}, respectively.
Control questions were used by $\jckpctcontrolquestions$\% of the publications to gauge annotator performance and task quality.
Overall, $\jckpctqualityestimation$\% of works mention at least one way of estimating quality.

\subsection{Quality Improvement}

Next, we analyze rectifying measures used to improve the data quality after it has been estimated in a previous step and deemed insufficient.
In most cases, incorrect or low-quality instances are corrected ($\jckpctfixbad$\%) or filtered out ($\jckpctdatafilter$\%). 
Of the $\jcknumdatafilter$ publications that mention filtering, $\jcknumagreementfilter$ report filtering based on agreement, $\jcknummanualfilter$ after manual inspection, and $\jcknumtimefilter$ based on unsound, improbably low annotation times.
$\jckpctautomaticchecks$\% of publications mentioned to have applied some kind of automatic checks to identify potential errors, such as spell checking or hand-crafted rules.
Sometimes, annotators were removed from the workforce if they repeatedly delivered sub-par quality ($\jckpctdeboard$\%).
Rarely were they given feedback by experts or the project managers ($\jckpctexpertfeedback$\%).
This number increases to $\jcknumexpertfeedbacknoexpert{}\%$ when excluding datasets only annotated by experts.
Overall, we do not see much usage of rectifying measures; only $\jckpctrectifyingmeasures$\% of publications using human annotation report at least one.

\subsection{Adjudication}

Similarly to \citet{sabouCorpusAnnotationCrowdsourcing2014}, we find that majority voting was most often used to adjudicate labels ($\jckpctadjmajorityvoting$\%).
In a few cases, publications reported that in addition to majority voting, ties were broken by consulting additional workers or experts ($\jckpctadjbreakties$\%).
The second most common way of adjudication was manual curation ($\jckpctadjmanualcuration$\%).
Overall, we find that in $\jckpctadjquestionmark$\% of labeling datasets, adjudication methods were not reported clearly or at all. 
This leaves the reader to guess, which is concerning.

We only found two publications that used \emph{Dawid-Skene}~\citep{dawidMaximumLikelihoodEstimation1979} and one that used \emph{MACE}~\citep{hovyLearningWhomTrust2013}.
The latter was just used to filter out spammers during annotation and not for adjudication itself.
One publication mentioned trying out probabilistic aggregation, yet they report that just using majority voting yielded better results for them.
Some works also mentioned aggregation based on annotator confidence and skill, but no details were given describing the exact procedure used.

The fact that majority voting is by far the most frequently used method is interesting, as aggregation is a quite well-researched topic in the crowdsourcing research community~\citep{sheshadriSQUAREBenchmarkResearch2013}.
It has also been shown that using more intricate methods can create higher-quality gold standards~\cite{paunComparingBayesianModels2018, simpsonBayesianApproachSequence2019}.

\subsection{Error Rate}
\label{sec:analysis_error_rate}

While it is often assumed that (research) datasets represent a gold standard and do not contain errors, this is often not the case~\citep[e.g.,][]{northcuttPervasiveLabelErrors2021, klieAnnotationErrorDetection2023}.
To estimate the overall correctness of the dataset, its annotation error rate should be computed after adjudication is completed.
Computing the error rate is typically done by randomly sampling a subset and marking instances as correct or incorrect.
From our analysis, only a few publications ($\jckpcterrorrate$\% of all having human annotation) estimated and reported an error rate.
The average error rate reported is $\jcknumerrorratemean$\%, and its median is  $\jcknumerrorratemedian$\%. 

\paragraph{Sample Size}

From the dataset we analyzed, $\jcknumerrorrateonlysubset$ out of $\jcknumerrorratenumreported$ error rates were computed by inspecting only a subset of the data.
The inspected subset needs to be of sufficient size for the estimate to be reliable.
If it is too small, the estimate has large error margins and hence low statistical power, potentially leading to over-optimistic or incorrect conclusions~\citep{buttonPowerFailureWhy2013, passonneauBenefitsModelAnnotation2014}.

For instance, it was found that \mbox{TACRED}~\citep{zhangPositionawareAttentionSupervised2017}, a dataset for relation classification, contains a large fraction of incorrect labels.
During the dataset creation, 25\% of the annotations were validated by crowdworkers; after adjudication, the authors finally inspected a sample of 300 instances and estimated an error rate of around $6.7\%$.
It was then subsequently discovered that the dataset contains significantly more errors.
First, it was claimed to be around $50\%$ by \citet{altTACREDRevisitedThorough2020}, who only analyzed a smaller and biased sample.
\citet{stoicaReTACREDAddressingShortcomings2021} finally inspected all samples and found an error rate of $23.9\%$.
This shows the importance of manual inspection of large enough sample sizes.

In the publications inspected, we did not find any work that based their choice of sample size on a statistical footing or gave reasoning for selecting that specific value.
In most cases, pretty numbers were chosen without rationale (e.g., round numbers like 100 or 200 were picked often), or a percentage of the total size (e.g., 5\%) was used.
The mean sample size is $\jcknumerrorratesamplesizemean$, while its median is $\jcknumerrorratesamplesizemedian$ (see \cref{fig:error_rate_sample_size}).

We also analyze the impact the sample size has on the estimate's reliability using confidence intervals and their interval half-widths.
The interval half-width measures the margin of error associated with the confidence interval.
 It is computed as the largest distance between the point estimate of the error rate and its endpoints.
The confidence interval for an estimated error rate $\hat{r}$ is then given as $[ \hat{r} - h, \hat{r} + h]$.
If $h$ is relatively large, e.g., $0.05$, then the error rate is with high probability within $\pm$ five percentage points.
This is quite a large margin, especially for error rates, as $\hat{r}$ is usually small there and (hopefully) close to zero.

To compute the margin of error, we model estimating the error rate as sampling with replacement\footnote{The sample size is usually much smaller than the dataset size, which is why we can approximate the hypergeometric distribution (sampling without replacement) with the binomial distribution for simplicity.} where annotators randomly inspect a subset of instances and mark them as either correct or incorrect.
For each mention of error rates in our analyzed publications, we then compute a 95\% binomial exact confidence interval for each estimate and its half-width $h$.

The half-widths for each estimate are plotted in \cref{fig:error_rate_sample_size}.
For almost all estimates, the resulting confidence intervals are very wide, rendering a given point estimate statistically unreliable.
When choosing a different sample to inspect and mark, the error would fluctuate by a large margin and has thereby only limited explanatory power.
We suggest inspecting at least 500 instances\footnote{Assuming a binomial model with a true error rate of 5\%, a sample size of $456$ yields a 95\%  CI with $h \approx 0.02$} or the whole dataset, whichever is smaller, for a more sound estimate.
Note that calculating the sample size that way is an optimistic estimate, as it assumes independent and identically distributed instances, which is often not the case.
Also, giving a confidence interval when stating the error rate is recommended.
This can either be done by computing a binomial/hypergeometric confidence interval or using techniques like bootstrapping.
Otherwise, giving a point estimate implies precision which it has not, especially when giving several decimal places.

\begin{figure}[bh]
    \centering
    \includegraphics[width=\columnwidth ]{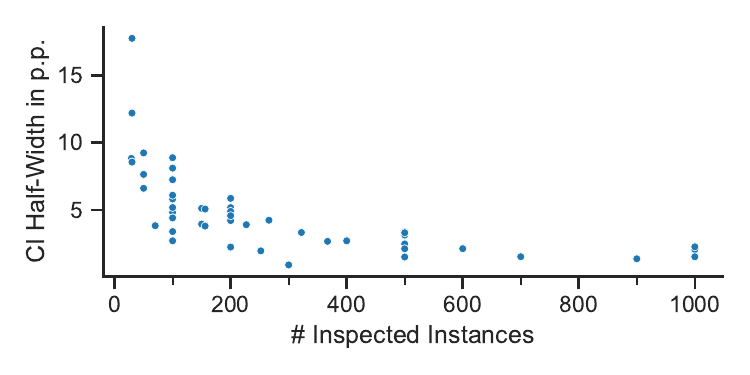}
    \caption{Number of inspected instances vs. the resulting confidence interval (CI) half-width for a 95\% CI. It can be seen that overall, too few instances are inspected to estimate the error rate reliably, as they have a substantial margin of error. Four values above 1000 were filtered out to aid the visualization.}
    \label{fig:error_rate_sample_size}
\end{figure}

\subsection{Agreement}
\label{sec:analysis_agreement}

For every paper inspected, we annotated whether agreement measure usage was mentioned and recorded its type and value if it was. 
In most cases, agreement has been used to demonstrate the dataset quality after the annotation was completed.
Sometimes, agreement has also been used to either remove annotators or remove annotations.
We observe that $\jckpctagreement$\% of publications involving human annotators reported using at least one form of agreement.
Concerning the form of dataset creation, it is $\jckpctagreementlabeling$\% for labeling and $\jckpctagreementtp$\% for text production.
In addition, we find that $\jcknumagreementnohumanannotation$ publications that ---while not employing humans for the annotation itself--- leverage agreement during validation steps.
The usage statistics are depicted in \cref{fig:agreement_barchart}.
Overall, Cohen's and Fleiss's \textkappa{}, Krippendorff's \textalpha{}, and percent agreement were used the most, followed by $F_1$.
On average, each publication used $\jcknumagreementmeasurescountmean$ agreement measures with median $\jcknumagreementmeasurescountmedian$ (based on works that actually used at least one).
Percent agreement as the only measure was used in around $\jckpctagreementpercentonly$\% of all publications that use at least one method.
Only using percent agreement makes it difficult to estimate, interpret, and compare the dataset's quality, and its usage is therefore discouraged~\citep{krippendorffReliabilityContentAnalysis2004}.
In $\jcknumagreementunknownmethod$ cases, the used measures were not clearly named but only referenced as e.g. \textkappa{} or IAA (this is noted by a '?' in \cref{fig:agreement_barchart}).

Regarding the usage and reporting of agreement as an indicator for reliability, we found similar issues as described by~\citet{amideiAgreementOverratedPlea2019}.
Often, only the agreement value was stated without any interpretation or comment ($\jckpctagreementinterpnone$\%), which limits its explanatory power.
In many publications, the quality derived from the agreement was described with a freeform explanation, e.g., \textit{high}, \textit{fair}, \textit{substantial} ($\jckpctagreementinterpcustom$\%).
These frequently do not have a relation to the actual value, as, for example,  values $< 0.3$ were described as \textit{reasonable}.
Rarely was agreement compared to previous studies ($\jckpctagreementinterpprevious$\%) or an interpretation based on a range given by the literature was cited ($\jckpctagreementcomparetoliterature$\%).
This can partially be explained by only some datasets having a suitable predecessor as a reference.

In all cases, these ranges' limitations were not considered; for example, the ranges defined by \citet{landisMeasurementObserverAgreement1977} are based on binary classification. 
In contrast, several datasets introduced by the respective publications had more than two possible labels.
Also, several times, the stated ranges did not match the metric. 
For example, the ranges from \citet{landisMeasurementObserverAgreement1977} that apply to Cohen's \textkappa{} were instead used for Fleiss' \textkappa{}.
Several times, publications used pairwise agreement measures for more than two annotators and reported them pairwise. 
While that is valid in itself, additionally using multi-user measures like Fleiss' \textkappa{} or \textalpha{} is recommended.
We also found several cases where the usage of Cohen's \textkappa{} was reported, but more than two annotations per instance were obtained.
It is also discouraged to use correlation metrics as a measure of agreement.
We found $\jcknumagreementcorrelation$ ($\jckpctagreementcorrelation$\%) of publications that still reported its usage.
Last but not least, \textkappa{} or \textalpha{} was sometimes given in percent.
This can confuse the reader as these values are usually given as a value in $[-1, 1]$, and percent agreement is a distinct metric on its own.

\paragraph{Agreement values} We plot the agreement values for the most frequently used methods in \cref{fig:agreement_stripplot} together with the boundaries suggested by the literature (even though they are often subjective).
For Krippendorff's \textalpha{}, the values are rarely larger than $0.8$, which would indicate acceptable agreement according to \citet{krippendorffReliabilityContentAnalysis2004}.
Some are in the zone ($0.67 \leq \kappa \leq 0.8$), which indicates that the resulting annotations should only be used to draw tentative conclusions; the majority is even below that.
Many agreement values are on the lower side, hinting towards lower agreement or considerable ambiguity in the underlying task.

\paragraph{Agreement for Sequence Labeling} For sequence labeling datasets (e.g., Named Entity Recognition or Slot Filling), dataset creators either did not compute agreement or relied on per-token \textkappa{}, \textalpha{}, or classification metrics like precision, recall, and mainly F1.
\citet{brandsenCreatingDatasetNamed2020} argue that per-token agreement for sequence labeling comes with two issues.
First, annotators label sequences and not tokens, so the measure does not reflect the task well. 
Second, the data is imbalanced, as most tokens are labeled \texttt{O}, indicating no span.
Excluding this would result in an underestimate of the agreement.
They argue for using $F_1$ and averaging it between annotators. 
However, this is not chance-corrected and can only be used to compute pairwise agreement; averaging might lead to a loss of information.
Only a single paper~\citep{stabIdentifyingArgumentativeDiscourse2014} used Krippendorff's unitizing \textalpha{}\textsubscript{u} ~\citep{krippendorffReliabilityUnitizingContinuous1995} to compute agreement for sequence labeling.
\textalpha{}\textsubscript{u} in itself can directly support sequence labeling and is an excellent way to compute agreement in this setting.
We hence agree with \citet{meyerDKProAgreementOpensource2014} that unitizing agreement measures should be used if not as the only measure, then at least additionally.
Our conjecture for why unitizing measures are not used more often is that these are not very well-known, and their complex implementation hinders adoption.

\begin{figure}
    \centering
    \includegraphics[width=  \columnwidth]{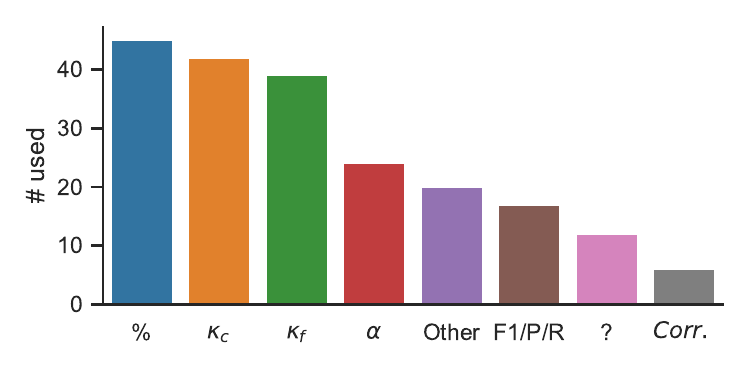}
    \caption{Distribution over counts of the agreement measures used. We count each method only once per publication, even if it has been used more than once. Overall, agreement measures were used in $\jcknumagreement$ publications involving human annotators. }
    \label{fig:agreement_barchart}
\end{figure}

\begin{figure}
    \centering
    \includegraphics[width=\columnwidth ]{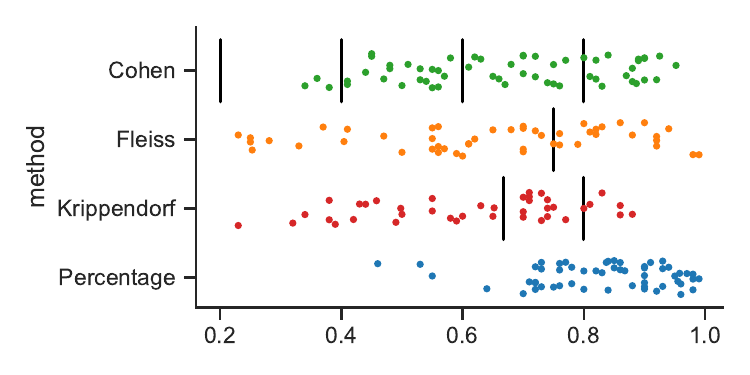}
    \caption{Agreement values for the papers inspected. Also shown are the ranges often used for interpreting these values.}
    \label{fig:agreement_stripplot}
\end{figure}

\paragraph{Sample Size}

Dataset creators sometimes decided only to have one annotation per instance for the majority of the dataset to save resources.
Then, only a subset was annotated multiple times to compute the agreement.
Similar to \citet{passonneauBenefitsModelAnnotation2014} and as described in \cref{sec:analysis_error_rate}, we note that having too small sample sizes is an issue as even a relatively relaxed 95\% confidence interval spans quite a wide range of values.
A sample size that is too small can cause estimates to vary by a large margin.
This might lead to a different interpretation based on a pre-determined, targeted agreement level or a range suggested by the literature.

Out of  $\jcknumagreementnumreported$ papers that reported agreement values, $\jcknumagreementfull$ have had the complete dataset annotated multiple times, $\jcknumagreementonlysubset$ were computed from a subset.
The mean sample size for the latter was $\jcknumagreementsamplesizemean$  with median $\jcknumagreementsamplesizemedian$.
$\jcknumagreementsamplesizeeqbelowtwohundred$ ($\jckpctagreementsamplesizeeqbelowtwohundred$\%) agreement values were computed on 200 instances or less, $\jcknumagreementsamplesizeeqbelowonehundred$ ($\jckpctagreementsamplesizeeqbelowonehundred$\%) even on less or equal than 100.

It is therefore recommended to 1) have large sample sizes to compute agreement on, ideally the complete dataset (which has the advantage of improved quality due to aggregation) and compute a confidence interval for the agreement value, e.g., by bootstrapping~\citep{efronBootstrapMethodsStandard1986, zapfMeasuringInterraterReliability2016}.
Computing the required sample size for a given precision and confidence level is not straightforward and depends on the metric~\citep{shoukriSampleSizeRequirements2004}.
For Cohen's \textkappa{}, an approximation is described by \citet{donnerGoodnessoffitApproachInference1992}; for \textalpha{}, it is given by \citet{krippendorffAgreementInformationReliability2011}.
As a rule of thumb that works for both \textkappa{} and \textalpha{}, given an expected/desired agreement value of $0.8$ with a precision of $h \pm 0.05$ and a confidence level of $95\%$, at least $\approx 500$ instances should be annotated.

While this is highly desirable, we notice that this comes with costs and additional effort.
We did not find a single report of confidence intervals for agreement values in the publications analyzed for this work.
As we do not have access to the raw, unadjudicated data used to compute the agreement value (which is needed for computing confidence intervals), we cannot easily conduct an analysis similar to the one for error rates in~\cref{sec:analysis_error_rate}.

\section{Recommendations}

Based on our analysis of \jcknumpubs{} papers published in top \ac{nlp} conferences as well as on our survey of the relevant literature, we derive the following recommendations and good practices for dataset creation quality control. 
A  case-by-case ranking of measures should be done based on the circumstances of the project.

\paragraph{Annotation Process}

\begin{itemize}
    \item Use an agile, iterative annotation process and annotate in batches~\citep{alexAgileCorpusAnnotation2010, pustejovskyNaturalLanguageAnnotation2013}. 
    \item Conduct pilot studies to validate the annotation setup before starting the actual annotation.
    \item Quality estimates after each batch should guide the improvement of guidelines and the scheme.
    \item Rectifying measures like corrective annotation, annotator retraining, or data filtering should be used to improve the overall data quality iteratively.
    \item Annotator feedback should be incorporated during a pilot study and annotation.
\end{itemize}

\paragraph{Annotator Management}

Workforce selection and annotator management are crucial for a successful annotation project.
Different annotator types can be viable depending on the task difficulty and the expertise or background knowledge required.
Datasets these days are most often annotated by crowdworkers.
A feasible alternative (even for tasks that usually require expert annotators) is hiring and training contractors via platforms like Upwork or Prolific.
This can open up better ways to collaborate while having similar costs. 

\begin{itemize}
    \item The choice of annotator type (expert/contractor/crowdworkers, \textellipsis) should be validated as part of a pilot study.
    \item Annotators should be paid properly and treated with respect.
    \item They should be trained before and during the annotation process for the best results, even experts.
    \item Annotator feedback should be used to fine-tune the guidelines, annotation scheme, or annotation editor and to spot errors or issues like low data quality.
    \item To select annotators, qualification tests are the recommended way; criteria like completed tasks or acceptance rate can be an addition, but should be rather lower than higher to not force workers into low-paying qualification jobs.
\end{itemize}

\paragraph{Quality Estimation}

Precise quality estimation is essential to steer the annotation process after each batch and before the final release of the dataset.

\begin{itemize}
    \item \Acl{iaa} can be used to determine whether the annotation process is overall reliable.
    \item In addition to agreement, manual inspection is recommended to validate annotations and estimate accuracy. This can be done by either the annotators themselves or experienced/expert annotators.
    \item Disagreements can be visualized using confusion matrices.
    \item An alternative to having annotators validate instances by marking them correct or incorrect is to have an additional task after the annotation/instance creation itself.
    \item Control instances can be injected into the data to annotate for measuring individual annotator performance and batch quality.
\end{itemize}

\paragraph{Agreement}

Agreement can be used to gauge how reliable the annotation process can be.
High agreement, however, does not automatically guarantee high-quality annotations and should be used together with other quality estimating and improving measures, like validation between annotation rounds or error rate estimation after adjudication.
Krippendorff's \textalpha{} can be used in almost all circumstances, even for sequence tagging in the form of unitized \textalpha{}~\citep{krippendorffReliabilityUnitizingContinuous1995}, continuous judgments, or with varying numbers of annotations per task and is therefore recommended.
The agreement value targeted should be chosen beforehand, either by pilot (expert) studies or previous annotation studies  annotating similar tasks. 
When the same number of annotators annotates each instance, Cohen's \textkappa{} for two annotators or Fleiss's \textkappa{} for multiple annotators can additionally be used, the latter only if annotators are randomly assigned to instances.
Percent agreement should rarely be used and never the only employed agreement measure.
Correlation coefficients like Pearson's $r$, Spearman's \textrho{}, or Kendall's \texttau{} should not be used to assess reliability.
Instead, Krippendorff's \textalpha{} or intraclass correlation is recommended as an alternative.

For a reliable estimate, agreement should be either computed on the whole dataset, or a sufficiently large ($\gtrapprox 500$ instances) subset should be annotated by multiple annotators.
Subset sample sizes should be statistically grounded, for instance, by computing them based on confidence intervals.
They should also be justified in the dataset description.
When using agreement, its usage should be reported in detail. 
The documentation should include which measures were used and why, how many judgments per instance were obtained, the background of the annotators, and the sample size used.
Agreement values require interpretation and should not stand alone. 
This can be done by defining a target agreement value, for instance, based on an expert study before the annotation itself, using a sufficiently high value like $0.9$, or comparing it to previous works.
Using thresholds from the literature like the ones from~\citet{landisMeasurementObserverAgreement1977} is not recommended, as these are arbitrary.
Confidence intervals should be employed to gauge the confidence of the agreement computation, whether they are reported as closed-form solutions given by the coefficient or via bootstrap.
More recommendations concerning agreement usage can also be found in the conclusion of \citet{lombardContentAnalysisMass2002}.

\paragraph{Quality Improvement}

Annotations are often not good enough at the beginning of an annotation project.
Therefore, estimating the quality and taking quality improvement steps is essential.
These can be, e.g., to correct low-quality instances or filter them out, improve guidelines and the annotation scheme, or train annotators.
Underperforming or adversarial annotators can be removed from the annotation project if required.

\paragraph{Adjudication}

Ideally, each instance should be annotated by multiple annotators in order to compute agreement and increase reliability via adjudication.
Majority voting is a strong baseline for aggregation; using more sophisticated approaches like \citet{dawidMaximumLikelihoodEstimation1979} or MACE~\citep{hovyLearningWhomTrust2013} might be worth trying, especially in settings where individual annotators are underperforming, or spammers are potentially prevalent.
Alternatively, expert curation or majority voting with experts breaking ties can be used to create a high-quality gold standard.
For reproducibility and better error analysis, it is suggested to not only publish the adjudicated corpus but also annotations by individual annotators.
These can then also be used to study and learn from the disagreement~\citep{umaLearningDisagreementSurvey2021}.

\paragraph{Error Rate Analysis} During and after the data has been annotated, it is crucial to have experts check the actual percentage of errors.
The sample size should be large enough to reach a high confidence estimate, which usually requires at least 500 instances (see \cref{sec:analysis_error_rate}) to inspect.
This sample size should be computed by considering the desired statistical guarantees, for instance, confidence level and estimated precision.

\paragraph{Reporting} 

We urge authors to accurately report on the annotation process when creating new datasets.
This includes, among others, annotator type and background, number of annotators, number of validators, dataset and subset sizes, agreement measures and values, adjudication methodology, and error rates.
In addition to that, we suggest augmenting the dataset documentation and reproducibility checklists (which are at the time of writing mainly concerned with model training and have only a few, if any, sections for dataset quality, see \cref{sec:background}), often required when submitting papers to conferences, with a section that is targeted with questions towards quality management good practices.
The checklist from \citet{kottnerGuidelinesReportingReliability2011} can be a good start for checking and guiding dataset creators toward the proper use of agreement.

\section{Conclusion}

High-quality datasets are essential for ---among others--- deducing new knowledge, for policy making, and to suggest appropriate revisions to existing theories.
They are also crucial for training correct and unbiased machine learning models.
If trained on datasets containing errors, inference can lead to wrong or biased predictions, which can cause material damage or even harm to other humans.
These potential issues are especially relevant with the recent, widespread adoption of conversational agents based on instruction-finetuned large language models.
Using datasets containing errors for evaluation can lead to incorrect estimates of task performance and, thus, to wrong conclusions when comparing models or approaches.

Quality management is an essential part of creating high-quality annotated datasets.
Therefore, we set out to better understand which  methods exist (RQ 1), which methods are actually applied  in practice (RQ 2), and how thorough (RQ 3).
For this, we surveyed the literature and inspected \jcknumpubs{} publications introducing new datasets from which \jcknumpubhashumans{} reported human annotation or validation, which we annotated for their quality management usage.

We answered our first research question by summarizing good practices for annotation quality management (\cref{sec:aqm}).
These are methods suggested in the literature or commonly used during dataset creation. 
Then, we used the dataset of annotated publications for their quality management to investigate which methods are used frequently and which are not.
Finally, we rated each publication for how well overall they conducted their quality management.
We found that, on the one hand, many works implement good practices very well. 
On the other hand, there are still issues that need to be improved on, for instance, better usage of agreement, annotator management, quality as well as error rate estimation, or reporting.
To be more precise, many papers used agreement without interpreting it, making it difficult to understand its implications.
Error rate and agreement were often computed on too small sample sizes, which renders the value imprecise and less expressive.
Frequently, annotation guidelines were not published, hindering reproducibility.

We conclude that many widely applicable techniques should be used more often or their use properly reported, especially iterative corpus creation as the annotation process of choice, pilot studies, validation, annotator training, qualification tests, control questions, annotation feedback, and debriefing, and maybe more complex adjudication.

We hope that our recommendations foster an adoption of good practices and an increase in dataset quality in the future.

\paragraph{Future Work}

In this paper, we analyzed $\jcknumpubs{}$ scientific publications introducing new datasets and annotated them for their annotation quality management.
We see several ways to build on this work.
First, while we already annotated a sizeable corpus of publications, using \textit{Papers With Code} introduced bias, limits analyzing quality management to what is reported in the paper and only contains a subset of dataset-introducing publications.
Therefore, we see the next step in a larger scale effort, ideally by directly asking authors to fill out a structured survey questioning them about their quality management.
While it might be difficult retroactively, it can be a good way for new datasets, especially when it is done as part of the publication and peer review process itself.
Second, it would be interesting to graph how quality management evolves over time and to analyze trends.
For instance, \citet{meyerDKProAgreementOpensource2014} state that agreement was not used very often in their small-scale analysis at the time, but we see that, on average, it is now used quite frequently.
Third, we only annotated which methods were used, but not what their actual, quantifiable impact was.
Hence, conducting such studies, similar to \citet{bayerlWhatDeterminesInterCoder2011} would be insightful, which analyze which factors contributed to higher agreement.
Fourth, as our work mainly focused on annotation and less on text production, we would like to see an extension in that direction.
Fifth, in this work, we focused on analyzing scientific publications concerning their quality management. 
We leave analyzing other aspects for future work, for instance, how well publications adhere to aspects checked for in dataset documentation or reproducibility checklists.
Sixth, it would be compelling to annotate the dataset by introducing publications on a large scale to alleviate the issues that our biased sampling might have caused.
This can then also be extended to other areas of machine learning, like computer vision.
Finally, we recommend that conference organizers and steering committees develop and adopt a dataset quality management checklist similar to existing ones and cover aspects like bias, intended use, or reproducibility.

\section{Limitations}
\label{sec:limitations}

In this work, one of our goals was to analyze how quality management of annotated datasets is done by inspecting and annotating the publications that describe their creation.
Our analysis already yields several relevant findings and common issues. 
We also were able to derive recommendations that future dataset creators can leverage for their own annotation projects.
However, we did not analyze the impact these practices have on the resulting dataset quality.
It is an interesting problem (but complex, as it requires manually analyzing not only the publications but also the datasets themselves) extension that we leave for future work.

We chose \pwc{} as the source of publications to annotate.
While our collection approach introduces bias and does not find all publications presenting new datasets, the papers annotated this way are for popular and frequently used datasets. 
Otherwise, they would not be listed in \pwc{}.
Our annotation still captures an important slice of quality management directly impacting research and state-of-the-art evaluation. 
However, a larger-scale annotation project would be the logical next step.

Our analysis relies on publications reporting their quality management. Hence, there might be a non-negligible underestimate of the numbers presented here.
New publications are inspired by how established datasets conduct their annotation process; therefore, even if good quality management is conducted, non-reporting is also an important issue that needs to be pointed out.

Our study is limited to primarily academic datasets and may have a blind spot in the industrial field, not only in terms of data but also in terms of methods.
However, this issue is difficult to alleviate, as industry datasets are often publicly unavailable.

The dataset is not intended to be used in machine learning, but is used to empirically underpin our survey.
Due to limited resources and the difficulty of the annotation task, each publication was only annotated by one annotator.
The impact on quality and consistency was reduced by repeatedly validating the annotations and using automatic rules to clean and improve them.
Ideally, more than one set of annotations would be available to compute agreement, adjudicate, and find errors, which we recommend for the next time.

For the overall rating, when conceiving the annotation guidelines and the scheme and during annotation, we tried our best to make it as objective as possible.
We still admit that the distinction between \textit{excellent} and \textit{sufficient} is relatively fluid.
However, we argue that our definition is relatively objective for \textit{subpar} quality management, which is the most relevant category for this work.
We were relatively lenient during annotation and assigned a better rating in case of doubt.
To further reduce the issue of subjectivity, we thought of alternatives like assigning scores based on the number of quality measures and their relative importance.
However, we ultimately abandoned this idea because not all works can use each measure, and we would have swapped one kind of subjectivity with another.

\section*{Acknowledgements}

We thank Falko Helm, Ivan Habernal, Ji-Ung Lee, Qian Ruan, Nils Dycke, Max Glockner, and our anonymous reviewers for the fruitful discussions and helpful feedback that improved this article.
This work has been funded by the German Research Foundation (DFG) as part of the Evidence (grant GU 798/27-1) and the PEER project (grant GU 798/28-1).

\clearpage

\begin{multicols}{2}
\bibliography{bibliography_clean, bibliography_dirty}
\end{multicols}

\clearpage

\appendix

\onecolumn

\section{Data Collection}
\label{sec:appendix_data_collection}

We use a snapshot of the \textit{Papers With Code}\footnote{\url{https://github.com/paperswithcode/paperswithcode-data}} data from the 26th of November, 2022.
From that, we select the \textit{text} datasets and match them against the ACL Anthology\footnote{\url{https://github.com/acl-org/acl-anthology}} with the commit \texttt{3e0966ac}.
While the ACL Anthology also contains backlinks to \textit{Papers With Code}, they were still very few ($\approx$100 datasets marked at the time of writing).
Hence, we opted to match them by title manually.

\begin{table*}[hb]
\centering
\begin{tabular}{@{}ll@{}}
\toprule
\textbf{File Name}                    & \textbf{md5}                     \\ \midrule
datasets.json.gz                      & 57193271ad26d827da3666e54e3c59dc \\
papers-with-abstracts.json.gz         & 4531a8b4bfbe449d2a9b87cc6a4869b5 \\
links-between-papers-and-code.json.gz & 424f1b2530184d3336cc497db2f965b2 \\ \bottomrule
\end{tabular}
\caption{File names and checksums for the \textit{Papers With Code} data.}
\end{table*}

\section{Guidelines}
\label{app:appendix_guidelines}

This annotation project aims to analyze how quality management is conducted in the wild.
In the following, we describe the different aspects we annotate.

\subsection{Manual Annotation}

We are mainly interested in analyzing works that use human annotators.
Therefore, we annotate whether a dataset involves humans as either annotators or validators.

\subsection{Task Type}

We see two broad categories of tasks that require different quality management methods.

\begin{description}
    \item[Annotation] This encompasses annotation projects where annotators provide labels, for instance, text classification, named entity recognition, annotating entailment for natural language inference, or selecting the right question from a given set for question answering.
    \item[Text Production] This encompasses annotation projects where annotators produce text. This can be, for instance, when writing surface forms that are later annotated. 
    Other tasks include summarization, question answering, dialogues, and natural language generation.
\end{description}

A dataset publication can use both task types, e.g., when creating questions and selecting the correct answer from a predefined pool or for natural language inference, where the clauses are first written and then labeled for their entailment.

\subsection{Annotators}

\begin{description}
    \item[Expert] We consider an annotator an expert if they annotate due to their domain knowledge or prior experience with the task.
    \item[Contractor] We consider an annotator a contractor if they are hired individually, for instance, student helpers or freelancers via platforms like Upwork or Prolific. The project managers usually know them by name and can directly interact with them. They can be managed on a more fine-grained level compared to crowdworkers.
    \item[Crowd] Crowdworkers are annotators who participate via platforms like Crowdflower or Amazon Mechanical Turk. Annotation is usually done in the form of microtasks. The annotators are relatively anonymous. There are often tens or hundreds of different annotators, each annotating only a small part of the overall data.
    \item[Volunteer] Volunteers are annotators who help for free and are not required to do so. This, for instance, excludes students who annotate as part of their coursework.
\end{description}

\subsection{Quality Management Methods}

\subsubsection{Annotation Process}

\begin{description}
    \item[Iterative Annotation Process] Mentions that an iterative feedback loop is used as the annotation process.
    \item[Pilot Study] It is mentioned that one or more pilot studies have been performed.
    \item[Data Filtering] Data is filtered before annotation via automatic or manual checks.
    \item[Validation] Mentions an explicit validation step. See \cref{app:guidelines_validation}.
    \item[Indirect Annotation] The annotation process has several steps, where the later ones indirectly validate earlier ones. 
\end{description}

\subsubsection{Annotator Management}

\begin{description}
    \item[Annotator Training] Training of annotators is mentioned.
    \item[Qualification Filter] It is mentioned that annotators are filtered out by criteria like native language, geographic location, previous acceptance rates, number of previously completed tasks, etc.
    \item[Qualification Test] It is mentioned that annotators had to take a qualification test before being allowed to participate in the annotation process itself.
    \item[Monetary Incentive] Give annotators additional payments if their quality is exceptional.
\end{description}

\subsubsection{Quality Estimation}

\begin{description}
    \item[Agreement] Uses at least one agreement measure. This must have been used for the annotation process or validation, not the pilot study. See \cref{app:guidelines_agreement}.
    \item [Error Rate] Computes the error rate for the final, adjudicated corpus. See \cref{app:guidelines_error_rate}.
    \item[Control Questions] Injects control questions for which the answer is known to estimate annotator and task performance.
\end{description}

\subsubsection{Rectifying Measures}

\begin{description}
    \item[Guideline Refinement] Mentions that guidelines and annotation schemes are refined.
    \item[Correction] Mentions that instances are improved and corrected.
    \item[Annotator Debriefing] Annotators give feedback to improve the annotation process.
    \item[Give Annotators Feedback] Annotators are given feedback to improve their annotation quality.
    \item[Agreement Filter] Instances are filtered out if agreement is too low.
    \item[Annotator Deboarding] Annotators are removed from the labor pool if their quality is deemed insufficient.
    \item[Manual Filter] Instances are filtered out manually if agreement is too low.
    \item[Time Filter] Instances are filtered out if annotators annotate improbably quickly.
    \item[Automatic Checks] Automatic checks are applied, for instance, spell checking or hand-crafted rules.
\end{description}

\subsection{Adjudication}

Adjudication describes the process of merging multiple annotations per instance into a single one.

\begin{description}
    \item[Majority Voting] The label assigned by at least half of the annotators is chosen. We also count adjudication as majority voting if all annotators must agree in the analysis, but label it as \textit{TotalAgreement}.
    \item[Manual Tie Breaking] A human annotator manually inspects instances without a majority and curates them. This adjudication method should be annotated together with \textit{Majority Voting}.
    \item[Dawid-Skene] This is an aggregation model that uses probabilistic graphical models to describe the expertise of the annotators.
    \item[MACE] This is an aggregation model that uses probabilistic graphical models to describe the expertise and likeliness of being a spammer of the annotators.
    \item[Manual Curation] A human annotator manually inspects and curates instances.
    \item[N/A] If there is only one annotation per instance or the task type is text production.
    \item[?] No mention of adjudication is found in the publication, but adjudication must have happened, e.g., because the publication mentioned more than one annotation per label.
\end{description}

If the task type is only text production, just enter \texttt{N/A} or leave the field empty; if annotation + text production, enter \texttt{?} or the mentioned one.
If you encounter new or different adjudication procedures, then please add them to the tagset.

\subsection{Guidelines available}

For reproducibility and to judge the quality of the annotation process, it is crucial that the guidelines are available. 
We consider guidelines available either in the publication, appendix, or supplementary material,  

\begin{itemize}
    \item a detailed annotation tagset/task/scheme description
    \item a screenshot of the annotation interface with a task description for the annotators
    \item or the guidelines itself
\end{itemize}
\noindent
are given.
We only check the external supplementary material if it is referred to in the publication.
In case the supplementary material is mentioned but not findable in the ACL anthology, we consider guidelines not to be available.

\subsection{Overall Judgement}

We assign an overall rating to each publication having human annotators based on their quality management conducted and reported. The grades are in three categories:

\begin{description}
    \item[Excellent] Does most of the following: uses the iterative annotation process, trains annotators, computes agreement and error rate, performs extensive validation, and does continuous human inspection.
    \item[Sufficient] Uses some of the recommended techniques, but not as extensive as excellent. Has at least some validation and manual inspection.
    \item[Subpar] No agreement, validation, manual inspection error rate, or other quality management performed and reported. The data quality, at most, relies on aggregation of multiple annotations. 
\end{description}

\subsection{Agreement}
\label{app:guidelines_agreement}

For each agreement value that is reported, create a new agreement annotation.
Agreement used in pilot studies should not be entered; we are only interested in values computed for the final dataset.

\subsubsection{Measure Name}

Enter the name of the measure. 
We are at least interested in the following:

\begin{itemize}
    \item Percent Agreement
    \item Cohen's $\kappa$
    \item Fleiss's $\kappa$
    \item Krippendorf's $\alpha$
    \item Krippendorf's $\alpha$ unitized
    \item Pearson's $r$
    \item Spearman's $\rho$
    \item Kendall's $\tau$
    \item Intraclass correlation coefficient    
    \item Precision
    \item Recall
    \item F1
\end{itemize}

Enter $?$ if it is unclear what the agreement measure is.
If you encounter new, different agreement measures, then please add them to the tagset.

\subsubsection{Value}

Enter the agreement value that is reported.
If no value is reported, but the use of agreement is, fill in as much as possible and enter $-1$.

\subsubsection{Inspection Size}

Enter the size of the subset that is used to compute agreement and the overall dataset size.
If the agreement is computed on the whole dataset, enter $0$ for both sample and total sizes.

\subsubsection{Interpretation}

We annotate the interpretation that is given together with the agreement value.
We are at least interested in the following works that give ranges for agreement measures and their interpretation.

\begin{description}
    \item[Landis] \textit{The Measurement of Observer Agreement for Categorical Data} by J. Richard Landis and Gary G. Koch, 1977.
    \item[Kripppendorf] \textit{Validity in Content Analysis} by Klaus Krippendorff, 1980.
\end{description}

If you encounter new, different works referenced that give interpretations, then please add them to the tagset.
We are also interested in

\begin{description}
    \item[Custom Interpretation] States that their agreement shows a certain level of quality, for instance, \textit{sufficient}, \textit{high}, \textit{good} without referencing a work from the literature.
    \item[Compares To Previous] Mentions a dataset that is similar to the one presented and compares its agreement to its predecessor.
\end{description}
    
\subsection{Validation}
\label{app:guidelines_validation}

We are interested in whether validation is done and who did the validation, if any. 

\subsection{Validators}

The labels for who is validating are the same as for annotators.

\subsubsection{Inspection Size}

Enter the size of the subset that is validated, as well as the overall dataset size.
If the complete dataset is validated, enter $0$ for both sample and total sizes.

\subsection{Error Rate}
\label{app:guidelines_error_rate}

The error rate is the number of incorrect instances divided by the total number of instances in the dataset.
We annotate it if it is computed on the adjudicated dataset.
It is usually computed on a subset of instances.

\subsubsection{Value}

Enter the error rate value that is reported.
If no value is reported, but the error rate is used, fill in as much as possible and enter $-1$.

\subsubsection{Inspection Size}

Enter the size of the subset that is used to compute the error rate as well as the overall dataset size.
If the error rate is computed on the whole dataset, enter $0$ for both sample and total sizes.

\section{Correlation}
\label{sec:appendix_correlation}

In the following, we give an example where correlation between ratings is high but agreement is low. We assume two annotators rating four items on a scale in $[1, 5]$:

\begin{center}
    \begin{tabular}{llrrrr}
        \toprule
        Item        & & a & b & c & d \\
        \midrule
        \multirow{ 2}{*}{Judge} & A     &     1 &  2 &  3 & 4 \\
        &  B    &     3 &  4 &  5 & 5 \\
        \bottomrule
    \end{tabular}
\end{center}

\noindent
The resulting correlation scores and are:

\begin{center}
    \begin{tabular}{rrrrrr}
        \toprule
        Pearson's \textrho{} & Spearman's \textrho{} & Kendall \texttau{} & ICC1 & ICC2 & ICC3 \\ 
        \midrule
        0.944 & 0.949 & 0.913 &  0.204 & 0.418 & 0.903 \\
        \bottomrule
    \end{tabular}
\end{center}

\noindent
It can be seen that standard correlation measures show very high correlation, while Intraclass Correlation scores are comparatively low.

\clearpage

\end{document}